\documentclass{WileyMSP-template}

\usepackage{graphicx}
\usepackage{epstopdf}
\usepackage{ragged2e}
\usepackage{amsmath, amsfonts, amssymb, amsthm}
\usepackage{cite}
\usepackage{float}
\usepackage{lineno}
\usepackage{xcolor}
\usepackage[font=sf]{caption}
\usepackage{listings}
\usepackage{xcolor}

\definecolor{codegray}{gray}{0.95}

\lstdefinestyle{arduino}{
  language=C++,
  backgroundcolor=\color{codegray},
  basicstyle=\ttfamily\footnotesize,
  keywordstyle=\color{blue},
  stringstyle=\color{orange},
  commentstyle=\color{gray}\itshape,
  numbers=left,
  numberstyle=\tiny,
  stepnumber=1,
  numbersep=5pt,
  showstringspaces=false,
  breaklines=true,
  frame=single,
  captionpos=b
}

\begin{document}

\pagestyle{fancy}

\rhead{\includegraphics[width=2.5cm]{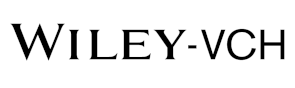}}

\title{Re-purposing a modular origami manipulator into an adaptive \\ physical computer for machine learning and robotic perception}

\maketitle

\justifying

\author{Jun Wang* and Suyi Li}

\begin{affiliations}
Department of Mechanical Engineering, Virginia Tech, Blacksburg, VA USA

*Corresponding author: \texttt{junw@vt.edu}

\end{affiliations}



\begin{abstract}
Physical computing has emerged as a powerful tool for performing intelligent tasks directly in the mechanical domain of functional materials and robots, reducing our reliance on the more traditional COMS computers. However, no systematic study explains how mechanical design can influence physical computing performance. This study sheds insights into this question by repurposing an origami-inspired modular robotic manipulator into an adaptive physical reservoir and systematically evaluating its computing capacity with different physical configurations, input setups, and computing tasks. By challenging this adaptive reservoir computer to complete the classical NARMA benchmark tasks, this study shows that its time series emulation performance directly correlates to the Peak Similarity Index (PSI), which quantifies the frequency spectrum correlation between the target output and reservoir dynamics. The adaptive reservoir also demonstrates perception capabilities, accurately extracting its payload weight and orientation information from the intrinsic dynamics. Importantly, such information extraction capability can be measured by the spatial correlation between nodal dynamics within the reservoir body. Finally, by integrating shape memory alloy (SMA) actuation, this study demonstrates how to exploit such computing power embodied in the physical body for practical, robotic operations. This study provides a strategic framework for harvesting computing power from soft robots and functional materials, demonstrating how design parameters and input selection can be configured based on computing task requirements. Extending this framework to bio-inspired adaptive materials, prosthetics, and self-adaptive soft robotic systems could enable next-generation embodied intelligence, where the physical structure can compute and interact with their digital counterparts.
\end{abstract}

\section{Introduction}

Over the past few years, we have witnessed the rapid emergence of embodied physical computing \cite{mengaldo2022concise, pitti2025informational}. The idea is to offload or ``outsource'' computation tasks from the electronic to the physical domain, significantly increasing the overall energy efficiency \cite{putra2024embodied}, parallelization \cite{yang2024efficient}, and resiliency against adversarial working conditions \cite{nygaard2021real}. To realize physical computing, researchers have explored two main strategies. One borrows the concept of algorithmic computing and uses bistable mechanisms as binary bits to construct logic gates for mathematical operations \cite{yang2025bistable, pal2023programmable} (e.g., using mechanical AND, OR, and NOR gates to construct adder circuits \cite{jiang2025modular}). Another strategy borrows the analog computing concept and uses dynamic and continuously variable signals to perform computation (e.g., using acoustic waves to solve differential equations \cite{zhang2023seismic} or using mechanical vibrations to perform machine learning tasks \cite{zonzini2022machine, cunha2023review}). One can purposefully build such ``algorithmic'' or ``analog'' physical computers from the ground up, using meta-material \cite{vidamour2023reconfigurable}, meta-surface \cite{zhou2022making}, or mechanical neural network architectures \cite{lee2022mechanical}. On the other hand, one can also re-purpose currently available physical systems --- such as soft robots \cite{wang2024proprioceptive, horii2021physical} and biological tissues \cite{yada2021physical, sumi2023biological} --- into computing devices, exploiting the computing capacity hidden behind their body dynamics. Implementing physical computation presents exciting potential to advance robotics, functional materials, wearable devices, smart infrastructures, and many other applications \cite{hauser2021physical}. 

Despite the rapid progress, physical computing is still a nascent topic with many ``big-picture'' questions to answer: What is the correlation between physical design and computing performance? How do we extract the most computing power from a physical structure? How can we apply physical computing to accomplish ``useful'' functions (beyond performing boolean logic and solving mathematical equations)? This study aims to provide insights into these questions using \textbf{P}hysical \textbf{R}eservoir \textbf{C}omputing (PRC) embodied in a modular and reconfigurable robotic manipulator.

\begin{figure}[t!]
    \includegraphics[scale = 1.0]{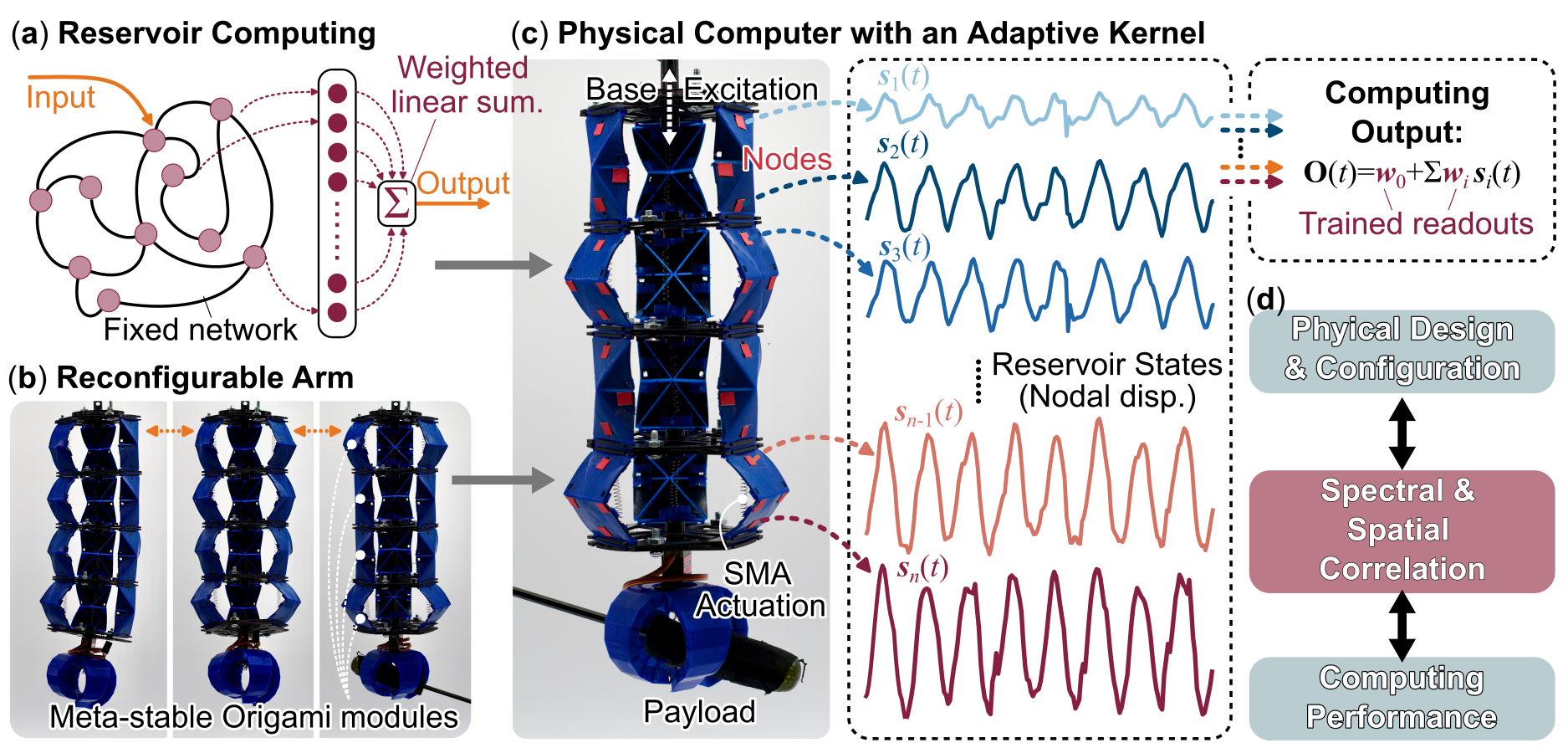}
        \centering
        \caption{The overview of this study. (a) A reservoir computer features a fixed neural network and a simple output layer of weighted linear summation. (b) In this study, we employ a reconfigurable robotic arm consisting of origami-inspired, meta-stable modules as the physical platform. (c) By applying the reservoir computing framework to this robotic arm, we creat a physical computer with an adaptive kernel, which can be dynamically excited with base excitation or embedded shape memory alloy (SMA) coil actuators. The dynamics of this adaptive kernel are represented by the displacements of the markers attached throughout its body, which function as the reservoir state vectors $s_i(t)$. The reservoir's computing output is simply a weighted linear summation of these nodal displacements $\mathbf{O}(t)=w_0+\sum w_i s_i(t)$, where the constant readout weights $w_i$ will be trained according to the tasks at hand. (d) In the big picture, this study reveals that the \textbf{spectral and spatial correlations} between the reservoir states and the targeted outcome is an informative hidden layer between the mechanical configuration of a physical computer and its computing performance.}
        \label{fig:bigpic}
\end{figure}

The physical reservoir computing framework utilizes a physical system's dynamic responses to perform machine learning tasks \cite{nakajima2018exploiting,nakajima2015information}. This approach is a subset of reservoir computing, which itself is a special type of recurrent neural network (RNN) characterized by its fixed computing kernel \cite{nakajima2020physical}. The core concept of physical reservoir computing is that a physical system capable of exhibiting a large degree of freedom and nonlinear dynamic responses can project an input excitation into a high-dimensional state space (i.e. reservoir states). These state-space vectors can then be ``read out'' to calculate the output through an analog and power-efficient weighted linear summation (\textbf{Figure \ref{fig:bigpic}}). Consequently, only the readout weights in the output layer need to be trained to achieve the desired input-output mapping for the machine-learning task at hand. In essence, physical reservoir computing treats the physical system as a fixed neural network, with training occurring solely in the output layer. Using this reservoir computing framework, various physical systems have successfully been transformed into physical computers across multiple domains, including mechanical \cite{he2024nonlinear}, electrical \cite{chen2024thin,moon2019temporal}, magnetic \cite{everschor2024topological}, photonic \cite{liang2023real}, and biological \cite{sumi2023biological} systems. 

In this study, we apply the reservoir computing framework to an origami-inspired, modular, and reconfigurable soft robotic manipulator, thus re-purposing it into a physically adaptive reservoir kernel for answering the aforementioned big-picture questions (Figure \ref{fig:bigpic}a-c).
Modular design is popular in robotics because of its scalability, reconfigurability, and versatility \cite{tejada2025review}. In particular, we make each module multi-stable using 3D-printed and pre-stressed origami panels so that each stable equilibrium has a unique shape and stiffness. Therefore, by adjusting the number of robotic modules and switching between their stable states, we can quickly adapt the physical reservoir's dynamic characteristics, creating the platform to uncover the relationship between the physical configurations and computing performance. Our extensive experiments show that the modular origami structure is a capable physical computer that can perform complex time-series emulation and information perception tasks in parallel. 

More importantly, via a thorough analysis of the adaptive kernel's dynamics at different configurations, we discover two metrics that can serve as the linkages between physical setups (i.e., mechanical design and constitutive properties) and computation performance (Figure \ref{fig:bigpic}d). These metrics are (1) \textbf{spectral correlation} between the reservoir dynamics and targeted output and (2) \textbf{spatial correlation} within the reservoir's state space. These two metrics are frequently used in nonlinear structural dynamics analysis \cite{fitch2025applying,vignesh2025review}, and we adapt them as design guidelines for physical computers: That is, they can inform the physical configuration and actuation of the reservoir to improve computing performance. 

Stepping further, using a robotic manipulator as a physical computer offers a natural pathway to applying its computing capacity to meaningful tasks. To this end, we equip the adaptive kernel with shape memory actuators and show that it could use its body dynamics to estimate the weight and orientation of the end payloads, as well as reconstruct the input command. These information can inform effective closed-loop control.
Overall, this study broadly proves that the mechanical design and input setup are closely related to physical computing performance. Given that the spectral and spatial correlations apply to any dynamic structures, the outcome of this study could be used to make many other soft robots and multi-functional materials more intelligent in the mechanical domain to meet demanding performance requirements.

\section{Setting up the Physical and Adaptive Computing Kernel}

\subsection{Origami-Inspired Metastable Module Fabrication}

\begin{figure}[b!]  
    \vspace{-0.2in}
    \includegraphics[scale=1.0]{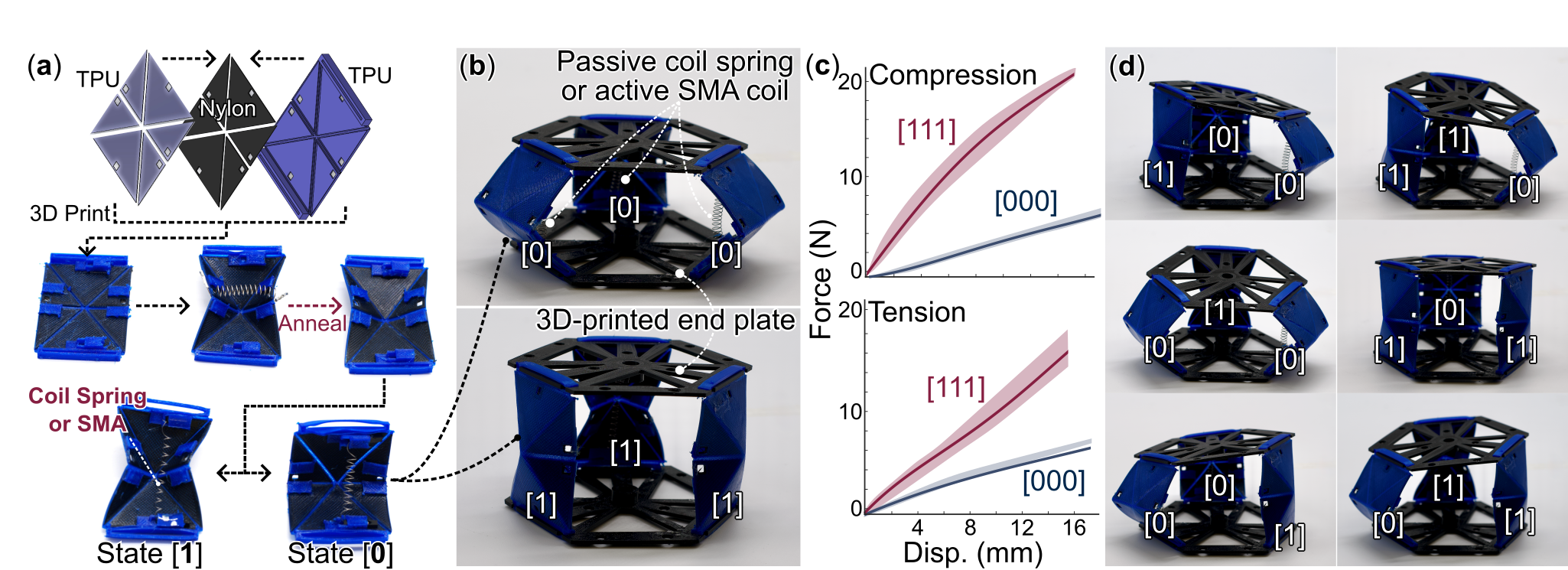}
        \centering
            \caption{Fabrication and configuration of the origami-inspired meta-stable modules. (a) To fabricate a bistable origami panel, we first 3D-printed it using flexible TPU and stiffer Nylon filaments and heat-treat it to reset its stress-free folding configurations. After that, we fit the origami panel with a coil spring or an SMA actuator coil, giving it the desired bistability featuring soft [0] and stiff [1] states. (b) We assembled three such panels and two 3D-printed end plates to complete the meta-stable module as the fundamental element in our adaptive physical computing kernel. Here, we focus on the most compliant [000] and stiff [111] states. (c) Tensile and compression testing reveal that the longitudinal stiffness ratio between the [111] and [000] states is $4.00\pm0.15$. The solid lines are averaged test results from 10 loading cycles, and the shaded bands are the corresponding standard deviation. (d) Close-up view of the six other metastable states from the same module.}
            \label{fig:setup}
\end{figure}

We design the robotic module to be lightweight, easy to fabricate, swappable, and with adjustable stiffness. This allows us to exploit its switchable properties to understand how physical reconfiguration influences computing capabilities.
Each module comprises two 3D-printed PLA base plates connected by three bistable origami panels (\textbf{Figure \ref{fig:setup}}). These origami panels are based on the classic Yoshimura pattern \cite{miura2010synthesis} and are 3D-printed using a dual-material FDM printer (Ultimaker S5). They feature a multi-layer construction: A 0.4mm thin base layer is first laid down using compliant TPU 95A filaments. Then, a middle layer of six 0.6mm thin Nylon triangles is added to increase the facet stiffness, creating the desirable folding kinematics. Finally, a 0.2mm thin TPU covers the whole panel to ensure proper bonding between different layers, even under large deformation. (Section 1 of the supplement material gives more fabrication details)

Once the 3D printing finishes, the origami panels are then manually folded inward along the creases, fixed by a coil spring, and then annealed at $110^\circ$ \text{C} for 10 minutes (Figure \ref{fig:setup}a). After the panels cool to room temperature, we remove the coil springs. This heat treatment eventually reset the stress-free resting configurations of the origami panels from flat to folded, thus conditioning them to exhibit multi-stability.
Finally, we re-attach the coil spring or a SMA coil actuator to the origami panel and create a bi-stable assembly: a flexible state [0] and a stiff state [1] (notice the coil spring is now in a different orientation than the previous heat treatment step). Therefore, once the three origami panels are attached to the base plates to create a robotic module, it possesses eight ($2^3$) different meta-stable configurations (Figure \ref{fig:setup}d).

Once the meta-stable modules are complete, we assemble them into a computing kernel simply using plastic screws through connection holes already printed in the base plates. 
The local stiffness and shape of these modules could be easily adapted by switching the origami panels between their [0] and [1] states. This study aims to examine the modular structure's adaptive computing capability, so we focus on the fully flexible [000] and the fully stiff [111] states. Compression and tension tests (Instron 6500 with 10 N load cell) show that the module in state [111] is approximately $4.00\pm0.15$ stiffer than in state [000] (Figure \ref{fig:setup}c). 



\subsection{Computation Tasks Overview}
\begin{figure}[t!]
    \includegraphics[scale=1.0]{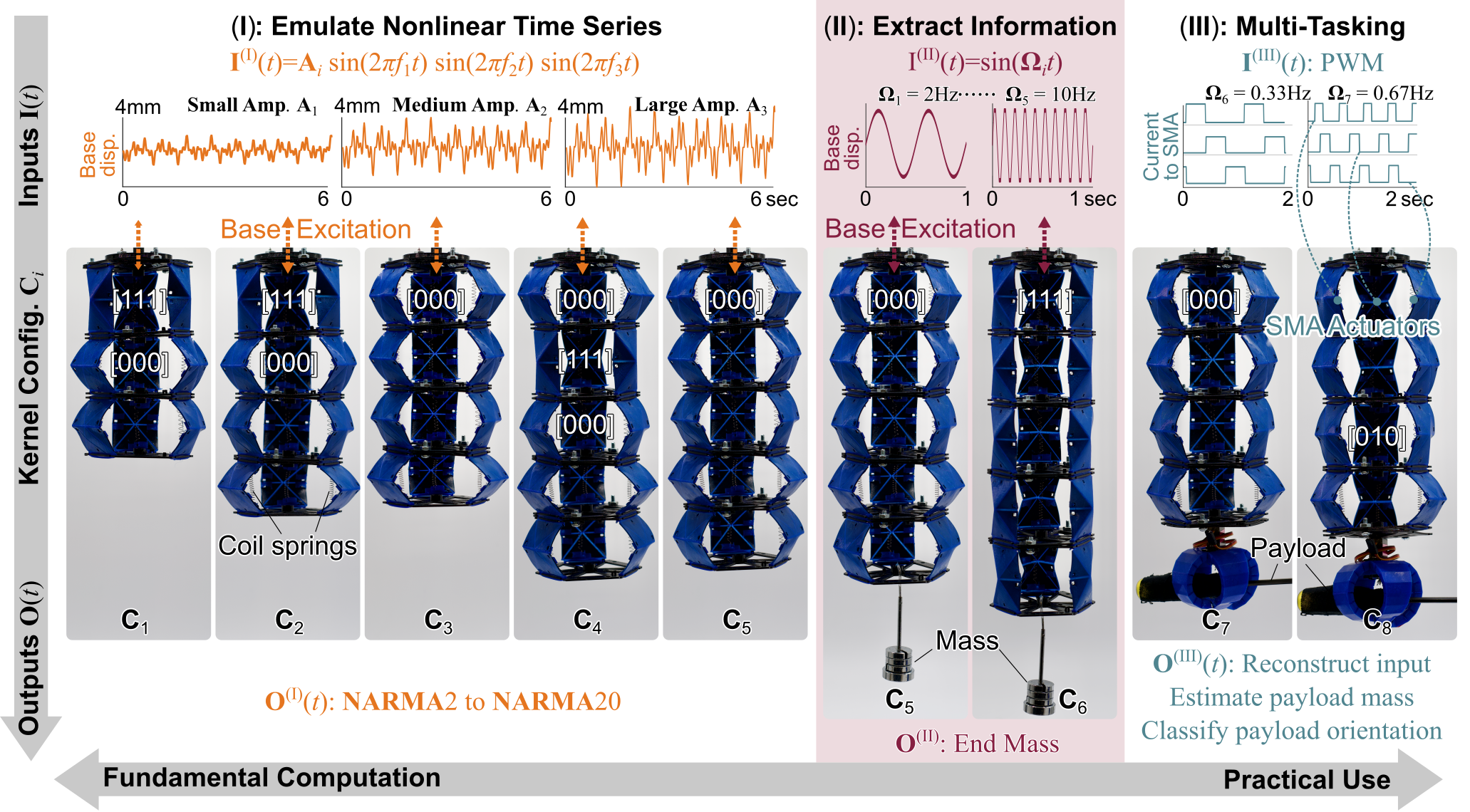}
        \Centering
        \caption{A summary of the computing tasks in this study, ranging from the more fundamental task (I) to the more practical multi-tasking (III). We investigated these three tasks using different physical configurations ($\mathbf{C}_1$ to $\mathbf{C}_8$) by adapting the number and stable states of origami modules. 
        In the time series emulation task (I), the input to the adaptive kernel is a base excitation $\mathbf{I}^{(\text{I})}(t)$ consisting of three harmonic signals; and the targeted output is a time series defined by nonlinear NARMA equations.
        In the information extraction task (II), we attached masses to the adaptive kernel's free end. The input is a simple harmonic base excitation with different frequencies, and the targeted output is the weight of these end mass.
        In the final multi-tasking (III), we replaced the embedded passive coil springs with active SMA coils and input Pulse Width Modulated (PWM) current to these SMA to excite (swing) the kernel. The output targets are to reconstruct the PWM input commands, classify the payload by estimating its weight, and determine the orientation of a payload.}
  \label{fig:Tasks}
\end{figure}

In this study, we test the adaptive physical kernel's capability to complete three computing tasks, aiming to answer the big-picture questions mentioned in the introduction.   These tasks cover a broad spectrum from the fundamental computing capacity (i.e., ``how much can the adaptive kernel compute?'') to the application of physical computing (i.e., ``how useful is the adaptive kernel?'') (\textbf{Figure \ref{fig:Tasks}}).

In the first Task (I), we challenge the adaptive kernel to complete the Normalized Auto-Regressive Moving Average (NARMA) task. This benchmark task quantifies the computational capacity of a neural network based on nonlinear time-series emulation \cite{wringe2024reservoir}. We test the adaptive kernel's NARMA performance using three base excitation magnitudes ($\mathbf{A}_1, \mathbf{A}_2, \mathbf{A}_3$) and five physical configurations ($\mathbf{C}_1, \ldots, \mathbf{C}_5$) to dive deep into the correlations between the computing performance and structural setup. 

In the second Task (II), we ask the adaptive kernel to estimate the weight of a random payload attached to its free end (Figure \ref{fig:Tasks}b).  This task only involves two physical configurations ($\mathbf{C}_5, \mathbf{C}_6$), but five different base excitation input frequencies ($\mathbf{\Omega}_1, \ldots, \mathbf{\Omega}_5$).  This setup lets us understand how to extract the most computing capacity from a given physical kernel. 

In the first two tasks, the adaptive kernel is attached to a large-stroke shaker (APS 113) that provides the input signals via base excitation. In the third and final Task (III), we embed shape memory alloy (SMA) coil actuators to transform the adaptive kernel into a functional robotic arm and challenge this robotic kernel to estimate the weight and orientation of its payloads as well as to reconstruct the input commands. Here, proper input frequency settings ($\mathbf{\Omega}_6, \mathbf{\Omega}_7$) and physical configurations ($\mathbf{C}_7, \mathbf{C}_8$) are also crucial for achieving satisfactory performance. 

Although the experimental setups and adaptive kernel configurations differ across these three tasks, the core computing framework is the same. The modular structure serves as the physical reservoir or mechanical neural network, mapping low-dimensional dynamic input into a high-dimensional state space, which is represented by the nodal (marker) displacements ($s_i(t)$) along the structure. In this study, we measure these nodal displacements by processing the camera-captured videos offline (Sonic a7C II camera with MATLAB image processing toolbox, Figure \ref{fig:bigpic}c). The computing output $\mathbf{O}(t)$ is simply a weighted linear summation of these state space vectors in that: 

\begin{equation}
    \mathbf{O}(t) = w_0 + \sum_{i=1}^{n} w_i s_i(t),
\end{equation}

\noindent where $s_i(t)$ is the displacement of the $i_{th}$ node, and $w_i$ is the readout weight (in this study, $n=40$).  The core principle of reservoir computing states that, with proper training, one can find an optimal set of readout weights so that the reservoir output matches the targeted output $\mathbf{O}(t)\approx\hat{y}(t)$, thus completing a machine learning task.

\section{Results and Insights}
\subsection{Task (I): Adapting the Physical Kernel for Optimal Time Series Emulation}


This first task aims to uncover how to configure our adaptive kernel to maximize its computing performance. To this end, we use the widely adopted NARMA emulation task to benchmark the computing output \cite{wringe2024reservoir}. In the experiment, the modular structure is attached to a vertically vibrating, long-stroke shaker that provides the input signal as the base excitation (supplement video S1). This input signal is a product of three harmonic streams and an amplitude scaling factor $A_n$:
\begin{equation}
    \mathbf{I}^{(\text{I})}_i(t)= \mathbf{A}_i \sin 2\pi f_1 t \sin 2\pi f_2t \sin 2\pi f_3t. \label{eq:NARMAin}
\end{equation}
 
\begin{figure}[ht!]
  \centering
  \includegraphics[scale=0.9]{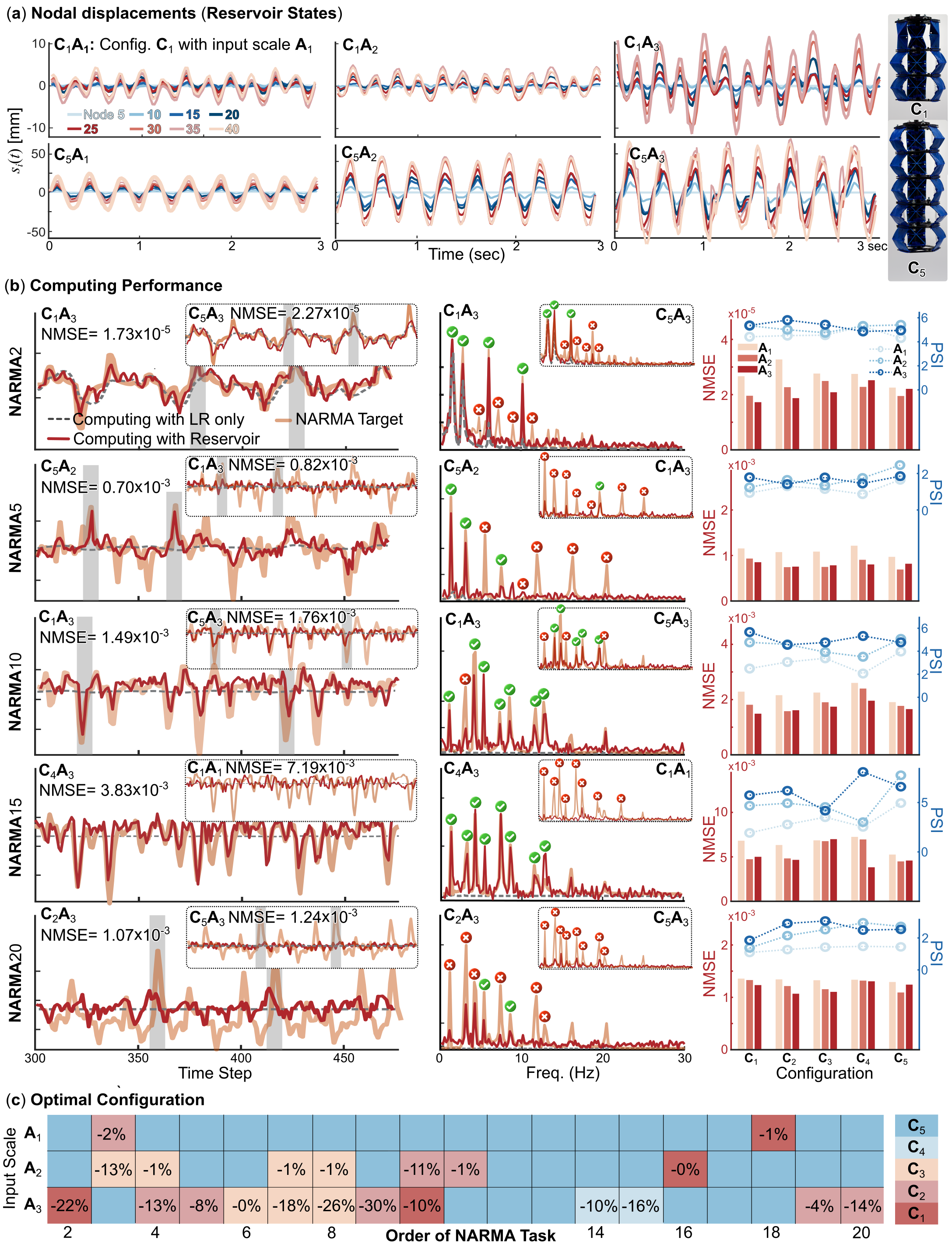}
  \caption{The adaptive kernel's computing performance using the NARMA benchmark. The reservoir dynamics differs, with different physical configurations $\mathbf{C}_m$ and base excitation magnitude $\mathbf{A}_n$. 
  (a) The nodal displacements of \( \mathbf{C}_1 \) and \( \mathbf{C}_5 \) under different input magnitudes. 
  (b) From the left to the right columns: The temporal outcome, spectral comparison, and NMSE/PSI summary of for NARMA2, NARMA5, NARMA10, NARMA15, and NARMA20 tasks. 
  (c) The adaptive kernel's optimal configuration corresponding to different NARMA tasks and input magnitudes. The values on this color map is the percentage reduction of NMSE error compared to the $\mathbf{C}_5$ configuration performance at that input magnitude.}
  \label{fig:RC_NARMA}
\end{figure}

The frequencies of the three harmonics are 2.11, 3.73, and 4.33Hz, respectively. All nodal displacements (or marker displacements) $s_i(t)$ are captured by a camera at 60 frames per second and extracted using the MATLAB image processing toolbox. \textbf{Figure \ref{fig:RC_NARMA}}(a) gives examples of these displacement fields.

Essentially, the nonlinear mapping between a one-dimensional input stream $\mathbf{I}^{(\text{I})}(t)$ and the corresponding displacement fields $s_i(t)$ captures the complex dynamics of the adaptive kernel. If this dynamics is ``rich'' enough, it can emulate the temporal response of a high-order nonlinear dynamic system defined by the NARMA task. In other words, via proper training at the output readout layer (explained in the supplement materials Section 2), one can find a set of readout weights $w_i$ so that the adaptive kernel's output $\mathbf{O}^{(\text{I})}(t) \left(=w_0+\sum w_i s_i(t)\right)$ can emulate the response of a nonlinear system $\hat{y}(t)$ in that
\begin{equation}
\mathbf{O}^{(\text{I})}(t)\approx \hat{y}(t), \text{ where } \hat{{y}}(t + 1) = \alpha \hat{{y}}(t) + \left( 5 \beta \hat{{y}}(t) \sum_{j=0}^{N-1} \textbf{I}^{(\text{I})}(t - j) \right) + \gamma \textbf{I}^{(\text{I})}(t - N + 1) \textbf{I}^{(\text{I})}(t) + \delta, \label{eq:NARMAout}
\end{equation}

Here, the parameter $N$ describes the complexity of targeted nonlinearity and, thus, the difficulty levels of the NARMA emulation task. Generally speaking, a reservoir that can finish higher-order NARMA tasks can accomplish more sophisticated machine learning \cite{wringe2024reservoir}. This study focuses on $N=2\ldots20$.

The input and output equations (\ref{eq:NARMAin}) and (\ref{eq:NARMAout}) indicate that the NARMA task challenges the adaptive kernel to generate higher-order harmonics output from a relatively simple input. This behavior is very similar to ``super-harmonics,'' a well-studied concept in nonlinear structural vibrations. Such a parallel offers us a unique opportunity to adopt knowledge from the discipline of structural vibration to physical reservoir computing here. For example, most vibration experiments concluded that generating strongly nonlinear super-harmonic dynamics requires two conditions. One is the intrinsic mechanical property: the structure must exhibit nonlinear and non-uniform stiffness. The other is extrinsic input: the excitation magnitude must be large enough to evoke such nonlinearity. 

Therefore, we adjust two types of ``physical hyper-parameters'' in this study to explore their relationship with computing capability. The first type corresponds to the intrinsic mechanical property, which includes the number of origami modules and their stable states. These factors dictate the physical dimensionality and nonlinearity. More specifically, we experimented with five different physical configurations with varying combinations of flexible and stiff modules. They are labeled as $\mathbf{C_1}$ to $\mathbf{C_5}$ as shown in Figure \ref{fig:Tasks}.
The second type of physical hyperparameter is the extrinsic input amplitude, and we used the shaker controller to input three levels of base excitation magnitude, labeled as $\mathbf{A_1}, \mathbf{A_2}, \mathbf{A_3}$ as defined in Eq.  (\ref{eq:NARMAin}). 
In each computing experiment, the overall reservoir setting is labeled as $ \mathbf{C_mA_n} $, where \( m \) represents the physical configuration and \( n \) is the input magnitude scale (e.g., $\mathbf{C_1A_1} $ means configuration 1, excited by the input magnitude 1). 

The first column of Figure \ref{fig:RC_NARMA}(b) summarizes the best temporal outcome from NARMA2, 5, 10, 15, and 20 tests and their corresponding $ \mathbf{C_mA_n} $ setup. For comparison, the smaller inserts at the upper-right corner show another outcome from a different, randomly selected setting. It is evident that, with the correct combination of physical configuration and input magnitude, the modular reservoir can closely emulate the targeted dynamics, even if the NARMA tasks ask for very high-order and complex features like the shaded area marked in Figure \ref{fig:RC_NARMA}(b). To quantify the NARMA computing performance, we adopt the \textbf{N}ormalized \textbf{M}ean \textbf{S}quare \textbf{E}rror (\textbf{NMSE}) between the reservoir and target outcomes (more on NMSE in supplement material Section 2). The errors from all computing experiments are summarized in the third column of Figure \ref{fig:RC_NARMA}(b).Interestingly, our adaptive reservoir outperforms a prior work using a significantly softer, silicon-based robotic arm, indicating that a lower material stiffness does not always lead to better reservoir computing outcomes (more on this in section 4 of the supplement materials). This is in contrast to the consensus in the reservoir computing community.


Another lesson from the discipline of nonlinear structural vibration is that \textit{spectral} analysis is as informative, if not more important, than the temporal response. Therefore, the second column of Figure \ref{fig:RC_NARMA}(b) shows the corresponding spectral content via fast Fourier analysis, which offers a much deeper insight into the computing performance of our adaptive kernel. More specifically, the targeted NARMA output contains several harmonic peaks: Some are at much higher frequencies than the input frequencies, and more difficult NARMA tests generally have more peaks. The adaptive kernel performs better if it can replicate these peaks via its inherent nonlinearity and superharmonic behaviors. In the spectral plot, we use green check marks to indicate a successfully reproduced peak and red cross marks to indicate a harmonic peak missed by the reservoir.

Therefore, we define a \textbf{P}eak \textbf{S}imilarity \textbf{I}ndex (\textbf{PSI}) to quantify the NARMA task performance in the spectral domain. The PSI is the ratio of the reservoir output's spectral peak magnitudes over the corresponding targeted peak magnitude based on NARMA. In this study, we focus on the first eight primary harmonic peaks. Therefore, the maximum PSI value is 8, meaning the adaptive kernel perfectly reproduces the first eight harmonic peaks required by the NARMA task. The minimum PSI is 0, meaning the reservoir fails to generate any desired harmonics (more details on PSI are in the supplement material Section 3). The third column of Figure \ref{fig:RC_NARMA}(b) summarizes the PSI from all computing tests, and the inverse correlation between Mean Square Error and Peak Similarity Index is evident.
 
\subsubsection*{Lessons Learned}


We can arrive at several intriguing conclusions by carefully assessing and comparing the computing performance from different physical settings.
First, a larger input magnitude helps to achieve better computing performance. The temporal error is lower (and spectral similarity is higher) when the input magnitude scale is larger at $\mathbf{A}_3$. This trend is particularly evident in the more difficult NARMA tests. The more difficult NARMA demands significantly more nonlinear dynamics (e.g., NARMA15 requires more harmonic spectral peaks than NARMA2). A larger input excitation can evoke such rich dynamics from the intrinsic mechanical nonlinearity. 

Second and most surprisingly, there is no optimal physical configuration (aka. there is no ``one size fits all'' solution to physical computer design, Figure \ref{fig:RC_NARMA}d). $\mathbf{C}_5$ is generally the favored configuration at a smaller input magnitude $\mathbf{A}_1$, likely because the smaller input does not invoke a significantly nonlinear response, and the additional modules in the $\mathbf{C}_5$ configuration can compensate for such a lack of nonlinearity. However, at the higher input $\mathbf{A}_3$ (with better overall NARMA scores), the optimal configuration changes between different NARMA tasks without apparent trends. 

These observations highlight the advantages of making the physical computer adaptive and modular. As different computing tasks can have widely different requirements, an adaptive physical kernel that can intentionally and intelligently reconfigure its physical architecture has a much broader chance of success.  

Finally, we conducted an additional experiment that only involved reconfiguring a 5-module adaptive kernel.  The outcome of this set of experiment are summarized in Section 4 of the supplement material. The lessons above still apply.

\subsection{Task (II): Extracting Payload Weight Information with Minimal Training Data}


While the previous task aims at fundamental computing capacity, this task focuses more on leveraging such computing power for more practical uses, like information perception and robotic control. In particular, we challenge the adaptive kernel to output the weight of a payload attached to its free end using the information encoded in its dynamic responses and minimal training data (Fig. \ref{fig:Tasks}).


\begin{figure}[ht!]
  \includegraphics[scale=1]{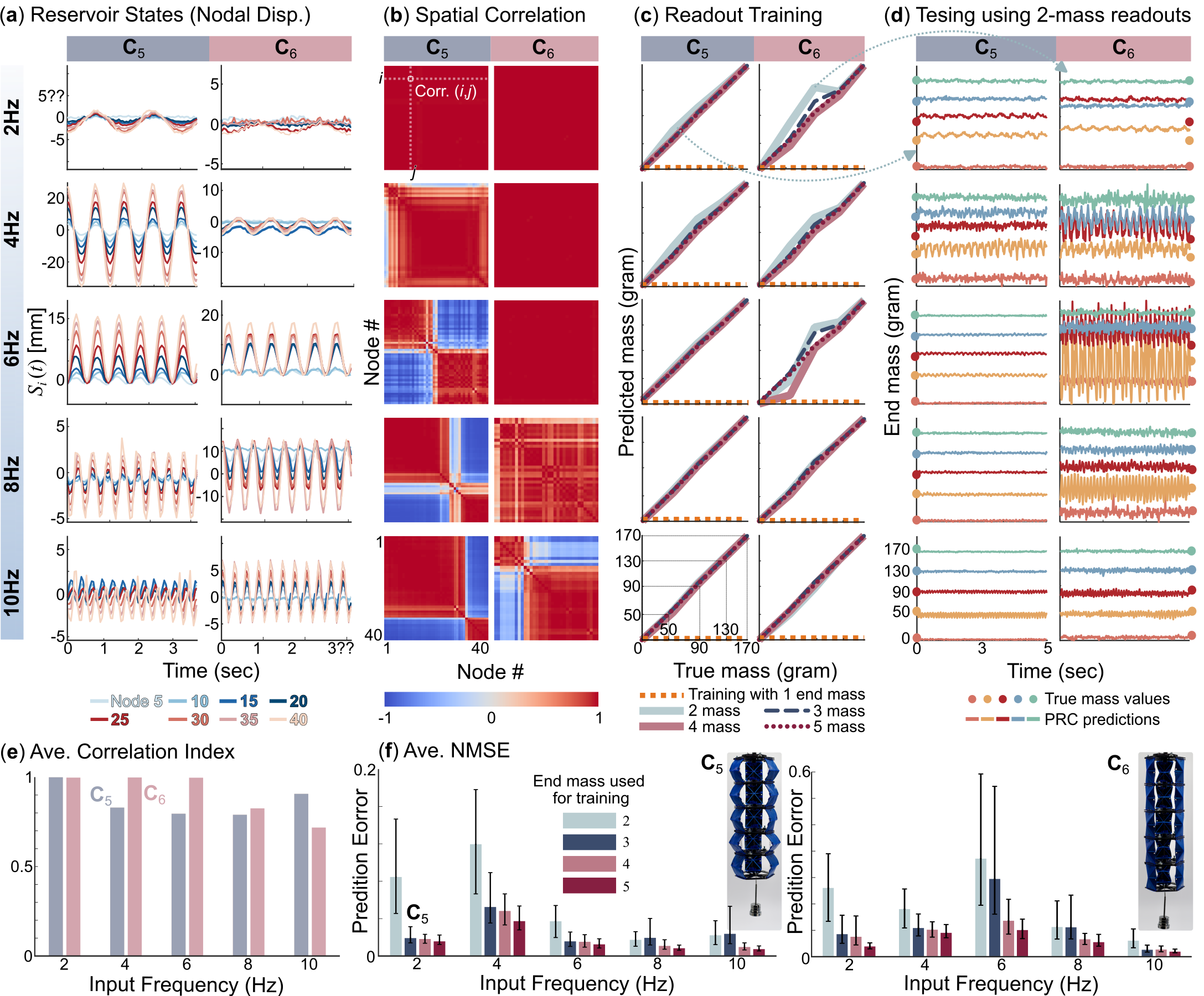}
  \caption{Extracting payload weight from body dynamics, using different kernel configurations and input conditions. (a) The adaptive kernel's body dynamics are represented by nodal displacements. (b) The correlation matrix between nodal displacements provides a quantitative measure of the ``richness'' of a physical body's dynamic characteristics. (c) The adaptive kernel estimates its payload mass using different amounts of readout training data.  If one uses the data from all five payloads for readout training, the kernel can output very accurate weight estimation. On the other hand, if one only uses the data from 1 payload, the kernel will fail the estimation task. (d) The reservoir's prediction using readout weights trained using data from two payloads. These results are accurate only when the spatial correlation between nodal displacements drops. (e) Averaged spatial correlation between different physical and input settings. (For visual clarity, only the lower left sub-plot's axes are labeled in b-d, and the same scale applies to all other sub-plots) (f) The overall estimation error in all testing cases with different readout training setups.}
  \label{fig:RC_payload}
\end{figure}

Unlike the NARMA tests that require strong nonlinearity, payload weight estimation is closer to a linear classification task. 
By exploiting our physical kernel's modular and adaptive nature, we can use this task to answer another ``big-picture'' question regarding physical computing: Once the physical design is fixed, how can we use minimal training data to achieve optimal computing performance? 
Ideally, the required training data is lower if the mapping from the payload to structural dynamics is a linear transformation, but a pure linear transformation might not be robust against the noises and uncertainties in the practical implementations \cite{inubushi2017reservoir}.   

To this end, we tested two configurations, $\mathbf{C}_5$ and $\mathbf{C}_6$, with five modules all in the compliant state [000] or all in the stiff state [111], respectively. The shaker input, in this case, is a simple harmonic stream $\mathbf{I}^{(\text{II})}(t)=A \sin \mathbf{\Omega} t$, where the harmonic frequency ranges from $\mathbf{\Omega}_1=2$ to $\mathbf{\Omega}_5=10$ Hz (Figure \ref{fig:Tasks}). 
The targeted output of this task is simply the payload weight, which has five distinct values: 0 grams (i.e., no payload), 50, 90, 130, and 170 grams. The readout weights $w_i$ are unique for each combination of input frequency and payload weight setting and must be trained separately (detailed readout weight training procedures are in the supplement materials section 5). 

First, we excite the adaptive kernel without payload at different base excitation frequencies to understand its fundamental dynamics. The corresponding state vectors $s_i(t)$ are recorded by the camera and summarized in \textbf{Figure \ref{fig:RC_payload}}(a). Here, we borrow another helpful concept from the structural vibration discipline: spatial correlation, which describes the similarity between the time responses from two points of a structure. A correlation of 1 means two signals are nearly the same, and a correlation of -1 implies the opposite. Figure \ref{fig:RC_payload}(b) summarizes the correlation between the different nodal displacements from Figure \ref{fig:RC_payload}(a). One can observe a ``threshold'' frequency, above which the correlation dropped significantly, meaning that the adaptive kernel's body dynamics become diverse and complex (supplement video S2). The magnitude of such a threshold frequency is inherently related to the intrinsic stiffness. The nodal displacement correlation started dropping above 4Hz with the softer $\mathbf{C}_5$ configuration and above 8 Hz input with the stiffer $\mathbf{C}_8$ configuration. This difference indicates that the correlation is directly related to the structure's primary resonance frequency. 


Then, we added payload masses to see if the adaptive reservoir could exploit its dynamics to estimate the weights. For each physical setup $\mathbf{C}_m\mathbf{\Omega}_n$, we attach the four masses one after another, thus obtaining five sets of temporal response data. This setup allows us to experiment using different amounts of data for readout training.
For example, we start by using 5 seconds of the nodal displacement data with two payloads (0  and 170 grams) for readout training. That is, denote $S_{0}(t)$ and $S_{170}(t)$ as the reservoir state space from 0-gram and 170-gram payloads, respectively. We can combine these two state spaces into a larger set $S(t)=[S_{0}(t) \; S_{170}(t)],$ and use the corresponding payload weight to define a step-wise function as the targeted output: 

\begin{equation}
    \hat{{y}}(t)= 
    \begin{cases}
    0 \quad &0\leq t \leq 5\\
    170 \quad &5\leq t \leq 10
    \end{cases}
\end{equation}

The readout weights $w_i$ can be obtained via linear regression so that the adaptive kernel's output $\mathbf{O}^{(\text{II})}(t)=w_i S(t)\approx \hat{y}(t)$(more details in section 5 of the supplement materials). Then, we apply this set of readout weights to the nodal displacements with other payloads to obtain a prediction of their weights. The results are summarized in Figure \ref{fig:RC_payload}(d).
With low input frequencies, when the structural dynamic correlation is high, the adaptive kernel performs poorly with a significant estimation error and oscillatory, unstable output. On the other hand, when the correlation matrix between nodal displacements starts to drop (or the body dynamics start to become richer and more complex), the modular kernel performs significantly better.

One can always use more data for readout training to improve prediction accuracy. Figure \ref{fig:RC_payload}(c) summarizes the adaptive kernel's prediction using 1 to 5 sets of nodal displacement for readout training (supplement material section 5). 
Suppose we use only one set of nodal displacement data (e.g., corresponding to 0-gram payload) to train a readout and apply this readout to the nodal displacements with other payloads. The adaptive kernel would fail the weight estimation task (dashed orange curves in Figure \ref{fig:RC_payload}c).
On the other end, we can achieve the best prediction by using all data with the five payloads for the readout training (red markers in Figure \ref{fig:RC_payload}(c)). This way, the prediction error is always lower than 2\% in the flexible $\mathbf{C}_5$ configuration and 10\% for the stiffer $\mathbf{C}_6$, regardless of the input frequency (dotted red curves in Figure \ref{fig:RC_payload}c). However, this is inefficient and wastes computing resources. 
%
%



\subsubsection*{Lessons Learned}

Our comparative tests show that complex and diverse body dynamics --- indicated by a low spatial correlation between different nodes --- are critical for reducing the required training data. Once the correlation is sufficiently low, one only needs the response data with two payloads for readout training. To reduce the correlation, one can reconfigure the adaptive kernel to lower its intrinsic stiffness (or resonance frequency).

\subsection{Task (III): Robotic Multi-Tasking with SMA-Activation}

\begin{figure}[ht!]
  \includegraphics[scale=1]{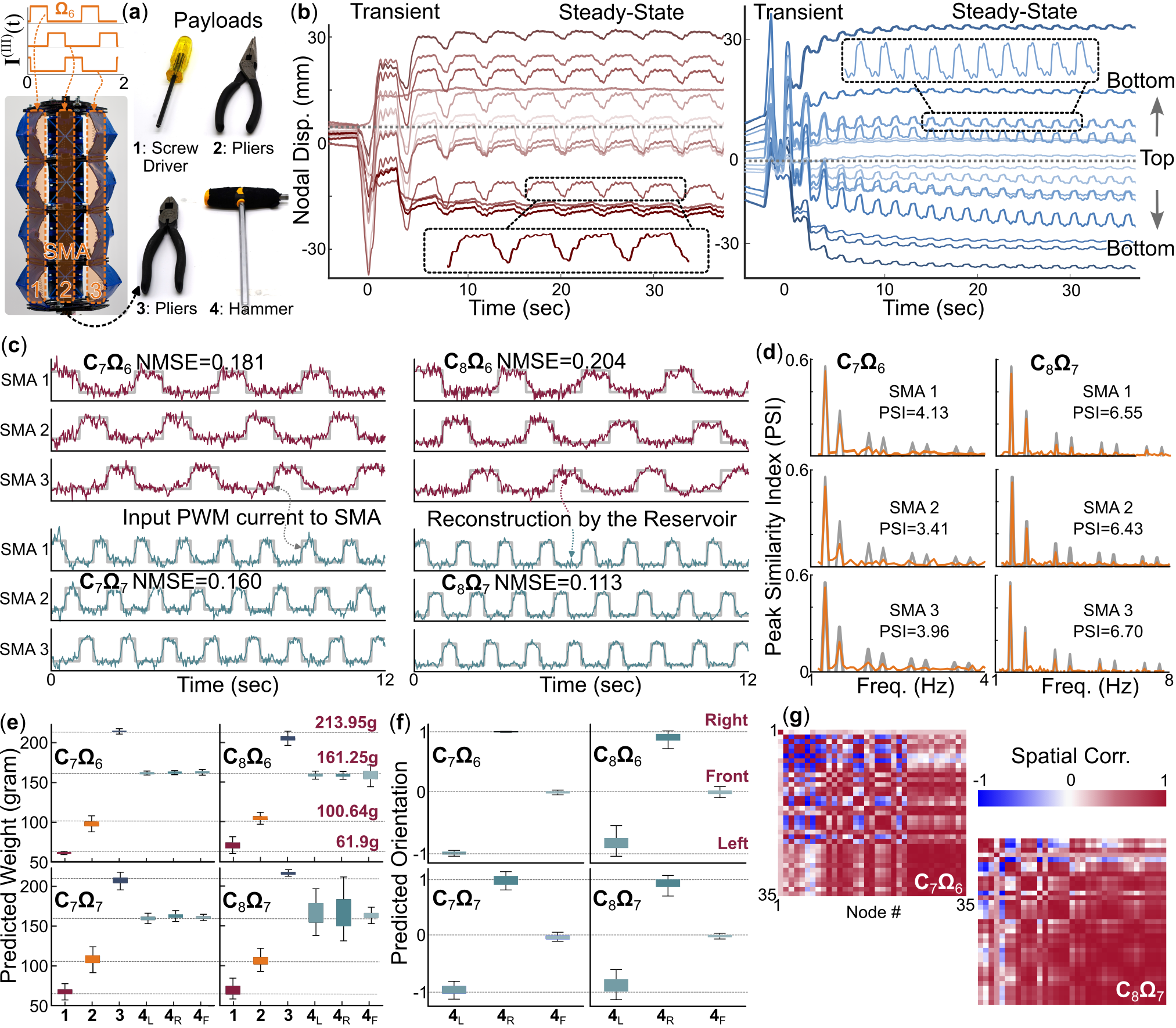}
  \centering
  \caption{Robotic multi-tasking (III). (a) The embedded shape memory alloy (SMA) coil actuator setup, the PWM input voltage command, and the payloads used in this study. (b) The nodal displacements under two different input frequencies. (c) The reconstructed SMA input commands from the adaptive reservoir based on four different setups. (d) The PSI of two example cases: $\mathbf{C}_7\mathbf{\Omega}_6$ and $\mathbf{C}_8\mathbf{\Omega}_7$. Similar to Task (I), higher PSI gives a better time-series emulation performance. (e) Payload weight estimation results, and (f) hammer orientation classification results. (g) Correlation matrix between different nodal displacements of the two example cases. Like Task (II), lower spatial correlation results in better information extraction performance.}
  \label{fig:SMA_payload}
\end{figure}

In the final task, we explore the potential of applying embodied physical computing for realistic robotic functions. To this end, we replace the passive coil springs with active shape memory alloy (SMA) coils in the origami modules, making the adaptive kernel into a robotic manipulator.  These SMAs can provide robotic actuation for object manipulation and structural support. Therefore, we attached different machine shop tools to the modular arm's free end as the payload, including a screwdriver, a hammer, and two pliers with different weights (\textbf{Figure \ref{fig:SMA_payload}}a). Here, the hammer has a significantly eccentric center of mass, which can substantially influence the manipulation outcome, so knowing its orientation is also helpful.  Therefore, we challenge the robotic kernel to achieve three computing sub-tasks.

The first sub-task is to emulate (or reconstruct) the input electric command to the SMA coils. This is a time-series emulation task similar to the NARMA Task (I); however, in this case, the targeted outcome $\hat{{y}}(t)$ is the voltage input to the SMA actuator in that $\hat{{y}}=\mathbf{I}^\text{(III)}(t)$. The second sub-task is to classify the end payloads by estimating their weights. If the end payload is the hammer, the robotic kernel must complete the third sub-task: differentiating the hammer's orientation. These two sub-tasks are similar to the payload weight estimation Task (II). It is worth highlighting that these three sub-tasks use the same nodal displacement data, so we are essentially re-purposing the robotic kernel into a multi-model sensor that simultaneously extracts complex information to inform effective control. 


We grouped the SMA coils into three columns and connected those within each column in series. This way, we can activate each column of SMA with a \textbf{P}ulse \textbf{W}idth \textbf{M}odulated voltage input (\textbf{PWM}), labeled as $\mathbf{I}^{(\text{III})}_1$, $\mathbf{I}^{(\text{III})}_2$, and $\mathbf{I}^{(\text{III})}_3$ respectively (Figure \ref{fig:SMA_payload}a). Such input can bend the robotic kernel for end payload manipulation while simultaneously generating the dynamic response required for reservoir computing. 
%


To investigate the influence of different dynamics on this task performance, we set the adaptive kernel into two configurations: $\textbf{C}_7$ and $\textbf{C}_8$. 
In $\textbf{C}_7$, all modules are configured in the entirely soft $[000]$ state. Meanwhile, in $\textbf{C}_8$, the modules are in the $[010]$ state, which creates asymmetric stiffness in the robotic arm so that it would bend even under uniform SMA actuation. In addition, we combined these two configurations with two PWM input patterns: $\mathbf{\Omega}_6$ and $\mathbf{\Omega}_7$. $\mathbf{\Omega}_6$ would joule heat the SMA coils for 0.1 seconds, followed by 0.2 seconds of cooling and relaxation, giving a 0.33Hz actuation frequency. Meanwhile, $\mathbf{\Omega}_7$ is 0.67Hz, giving 0.05 seconds of heating and 0.1 seconds of cooling (Figure \ref{fig:Tasks}). Consequently, we test four physical settings: $\mathbf{C}_7 \mathbf{\Omega}_6$, $\mathbf{C}_7\mathbf{\Omega}_7$, $\mathbf{C}_8 \mathbf{\Omega}6$, and $\mathbf{C}_8\mathbf{\Omega}_7$, as shown in Figure \ref{fig:SMA_payload}.


Regardless of the kernel configuration and PWM input setup, the modular robotic arm would exhibit random and significant transient dynamics due to the local buckling of some origami panels. After a few seconds, it settles into a steady state and small amplitude swing motion around an equilibrium shape (Figure \ref{fig:SMA_payload}b). Such swinging motions, caused by alternating actuation of the three columns of SMA coils, are recorded by the camera and serve as the training and testing dataset. Section 6 in the supplement material details the training and testing procedures.

Figure \ref{fig:SMA_payload}(c) shows the performance of the input command emulation tasks across different physical settings. The modular kernel's output can emulate (or reconstruct) the input command with reasonably high accuracy. In particular, configuration $\mathbf{C}_8 \mathbf{\Omega}_7$ achieved the best emulation performance. Moreover, higher-frequency actuation provides better input signal mapping. The increased stiffness and higher actuation frequencies enhanced the robotic arm’s dynamic synchronizing with the input command, leading to a higher spectral correlation between the input and output (i.e., higher peak similarity index, as shown in Figure \ref{fig:SMA_payload}d).

Figure \ref{fig:SMA_payload}(e) summarizes the results of payload weight estimation under the four physical settings. In all cases, reservoir outputs are well separated across repeated experiments, enabling clear differentiation among the four payloads. 
Interestingly, the optimal physical settings for payload estimation differ from those for input command emulation. 
In this case, the more flexible configuration ($\mathbf{C}_7$) exhibits more consistent performance with more minor errors away from the ground truth (MSE lower than 5\%). The prediction error is also significantly minor under the lower actuation frequency $\mathbf{\Omega}_6$. These observations suggest that the robotic kernel exhibits richer body dynamics under these conditions, exhibiting a lower spatial correlation between different nodal displacements (Figure \ref{fig:SMA_payload}g). 

Finally, suppose the end payload is identified as the hammer. In that case, we apply another set of trained readouts to the same set of nodal displacement data to predict its orientation (i.e., the hammerhead toward the left, right, or front). Figure \ref{fig:SMA_payload}(f) clearly illustrates that all physical settings allow the robotic arm kernel to distinguish the hammer’s orientation.



\subsubsection*{Lessons Learned} 
First, this multi-tasking experiment exemplifies the potential of embodied physical computing in soft robotic systems. By swinging the robotic arm using embedded SMA actuators, we can generate rich body dynamics that contain useful input commands and payload information. Physical reservoir computing is an efficient way to extract this information. 

Moreover, the results of this multi-tasking study point to a trade-off between different computing and information perception tasks. The optimal physical configuration and input setups depend on the computing task's nature. Flexible structures with lower actuation frequencies improve information extraction outcomes (i.e., classifying the payload weight and orientation). In comparison, stiffer configurations with higher actuation frequencies enhance time series emulation (i.e., input command reconstruction). This insight again highlights the benefit of using adaptive physical kernels for different embodied computation tasks so that one can reconfigure them accordingly.

\section{Discussion and Conclusion}

First, it is worth emphasizing that our adaptive kernel is repurposed from a fully functional soft robotic arm and is never specially designed for physical reservoir computing. Regardless, it shows impressive computing and machine-learning capability in the mechanical domain. In particular, this robotic arm's meta-stable and modular nature allows us to uncover the connection between physical configuration and computing performance. Therefore, one can apply the results of this study to many other reconfigurable soft robotic systems, uncovering their embodied computing power without sacrificing (or even improving) their original robotic capability.   

Secondly, because our adaptive kernel is repurposed from a robotic arm, we choose to use simple markers with image processing to extract its body dynamics --- in this way, we do not need to add any mechatronic components to the robotic arm. However, in a parallel proof-of-concept study, we show that it is also possible to add embedded sensors like strain gauges to measure body dynamics without a camera. In this way, one can achieve fully online physical computing and information extraction.

In summary, the adaptive and multi-stable modular arm is an ideal platform for exploring key questions in physical computing: What are the underlying connections between physical design and computing capability? How useful is this computing power for practical applications like robotic manipulation? By leveraging the modular arm's tunable stiffness and dynamic responses, we demonstrate that physical reservoir computing can be effectively integrated into soft robotics, bridging the gap between robotics and intelligent computing within the mechanical domain. Unlike traditional AI-driven control strategies, our system decentralizes computation to the physical body, enabling efficient embodied perception and decision-making. Through three tasks---adaptive computation (NARMA emulation), payload estimation, and multi-tasking via SMA actuation---we show that structural configuration, actuation frequency, and dynamic properties significantly influence computational performance. 

Critically, we show that \textbf{spectral correlation} (i.e., peak similarly index PSI) between the body dynamics and targeted output and the \textbf{spatial correlation} within the physical body are the key performance indicators that can inform the optimal configuration and re-configuration.  By merging materials science, robotics, and computational intelligence, this work lays the foundation for next-generation embodied AI systems, where robots use their physical structures as an intrinsic part of intelligent computation and control.

\medskip
\textbf{Supplement Information} \par 
Supporting Information is available from the Wiley Online Library or from the author.

\medskip
\textbf{Acknowledgements} \par 
Funding for this work was provided by NSF EFRI BEGIN OI \#2422340, NSF DCSD \#2328522, and Virginia Tech (via startup funding) 

\medskip

%

\bibliographystyle{MSP}
\bibliography{1_main}

\newpage

\section*{Supplement Materials}

\subsection*{Section 1. Origami Module Fabrication and Assembly}

\subsubsection*{CAD Modeling} 

\begin{figure}[ht!]
  \centering
  \includegraphics[width=\linewidth]{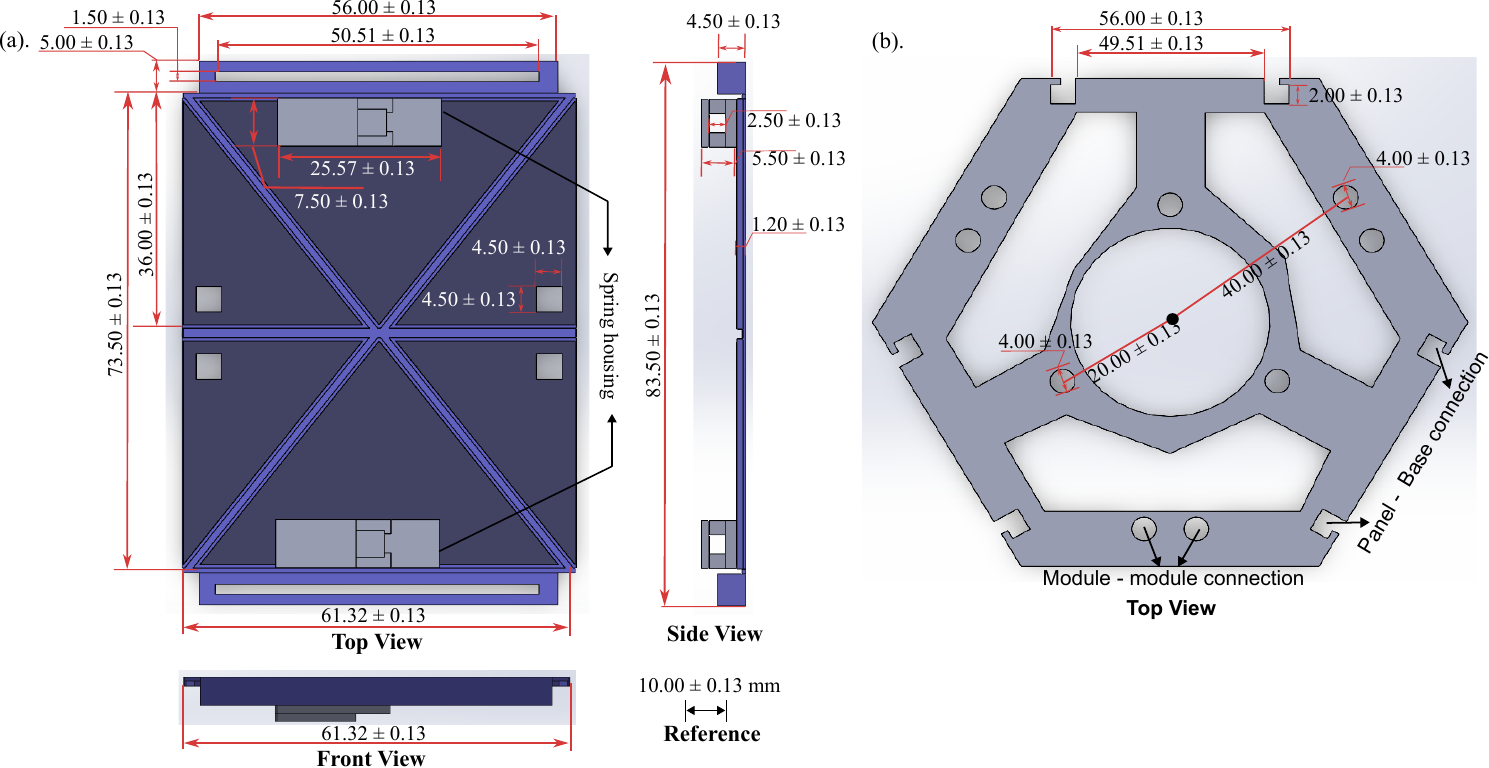}
  \caption{CAD drawings of the origami panel and base plate. (a) Dimensions of the origami panel in top, side, and front views. (b) Dimensions of the rigid base plate used for assembling three panels into a single module and connecting multiple modules.}
  \label{fig:S1_CAD}
\end{figure}

The 3D CAD model of the origami panel and the stiff connecting base plates is developed using \textsc{SolidWorks} 2024, as shown in \textbf{Figure \ref{fig:S1_CAD}}. The classical Yoshimura pattern inspires the crease design of the bistable origami panel---A rectangular shell composed of six triangular facets.  The panel comprises three layers: a 0.4mm thermoplastic polyurethane (TPU) 95A base layer, a 0.6mm stiff Nylon layer forming the six triangular facets, and a 0.2mm thin TPU top layer that covers and bonds the Nylon to the base. Nylon is selected as the printing material for the middle layer for its high stiffness, significantly enhancing the panel’s metastability. TPU 95A primarily serves as the crease material, ensuring sufficient flexibility for folding and reconfiguration.

Figure~\ref{fig:S1_CAD}(a) illustrates the top, side, and front views of the assembled origami panel. Several design modifications were made to adapt the structure for fabrication, distinguishing it from conventional paper-based Yoshimura models. First, a gap consisting solely of the TPU base layer (blue components in Figure~\ref{fig:S1_CAD}a) is introduced between the thick triangular facets (black regions formed by the TPU-Nylon-TPU sandwich structure) to enable proper folding, as it maintains negligible thickness and sharp fold lines. Second, three additional components are integrated to facilitate manufacturing. Hollow TPU cubes located at the top and bottom of the panel enable simple mechanical connection to the rigid plate by press-fitting into panel-base connectors positioned at the plate’s edge (Figure~\ref{fig:S1_CAD}b). Additionally, two TPU-printed housings are included to accommodate spring coils or shape memory alloy (SMA) actuators. Four square cutouts along the centerline are designated for placing the coil spring during the annealing process. The rigid base plate is designed as a 1mm-thick hollow hexagon, with panel-base connectors cut into three of its edges to allow simple attachment of the origami panels. Additionally, nine circular cutouts are incorporated into the plate to facilitate inter-module connections using screws.

\subsubsection*{3D Printing} 

In this study, we employed the widely used fused deposition modeling (FDM) technique to fabricate the origami panel and rigid base plate using a dual-material Ultimaker S5 3D printer. A layer-by-layer dual-material printing strategy was adopted for the origami panel following the methodology presented in our previous work~\cite{deshpande2024golden}.

To ensure acceptable printing resolution, the layer thickness for 3D printing was set to 0.1mm. First, the TPU 95A base layer (0.4mm thin) was printed using a zig-zag path and a printing core temperature of 225\degree C, providing a durable and flexible foundation. The Nylon triangular facets were then printed on top using a zig-zag pattern, with the core temperature set to 250\degree. Finally, a 0.2mm thin TPU top layer, printed under the same conditions as the base layer, was deposited to encapsulate and firmly bond the Nylon layer, ensuring structural integrity under large mechanical stress and deformation. The rigid base plate was printed solely with PLA material using a zig-zag infill pattern at a core temperature of 205\degree C.

\subsubsection*{Annealing} 

Before assembling the origami panel and rigid plate into the metastable module, the panel must be annealed to establish its bistable behavior with the aid of coil springs. As shown in Figure~\ref{fig:setup}(a), the printed panel is folded inward along its crease lines and temporarily held by a coil spring. This folded panel is annealed in a temperature-controlled oven (EasyComposites OV301) preheated to $110^\circ$C for 10 minutes to relieve internal stress in the bent facets. After annealing, the panel can cool to room temperature for 30 minutes to eliminate the residual stress. The horizontal spring is then removed.

In the final step, the head and tail of the coil spring or shape memory alloy (SMA) coil are inserted into the two 3D-printed housings on the origami panel. This setup enables the creation of a bistable structure with two distinct states: a flexible state ([0]) and a stiff state ([1]), as introduced in the main text.

\newpage

\subsection*{Section 2. NARMA Computing Setup and Readout Training}

\subsubsection*{More Details About NARMA} 

In Task (I), the meta-stable modular structure is utilized as a physical reservoir (adaptive kernel) to explore the optimal input and physical configurations for optimal computing performance. Its computational capability is evaluated through the \textbf{Normalized Auto-Regressive Moving Average (NARMA)} system emulation task. NARMA is a widely adopted benchmark for small-scale reservoir systems, as the difficulty of the task is well-characterized by the parameter \(N\), which serves as a proxy for the system’s memory capacity ~\cite{wringe2024reservoir}. The NARMA-\(N\) system can be expressed by the following equations. 

For \(N = 2\) (NARMA2) $\hat{y}^2$ could be expressed as :
\begin{equation}
    \hat{y}^2(t + 1) = 0.4\hat{y}^2(t) + 0.4 \cdot \hat{y}^2(t) \cdot \hat{y}^2(t - 1) + 0.6u(t)^3 + 0.1
\end{equation}

For \(N > 2\) (the general form of NARMA-$N$) $\hat{y}^N$ could be expressed as:
\begin{equation}
    \hat{y}^N(t + 1) = \alpha \hat{y}^N(t) + 5 \beta \hat{y}^N(t) \sum_{j=0}^{N-1} u(t - j) + \gamma u(t - N + 1) u(t) + \delta
\end{equation}
where \( u(t) \) is the input signal provided to the system, and \( \hat{y}^N(t) \) is the target output for the NARMA-\(N\) task. The parameters \( \alpha \), \( \beta \), \( \gamma \), and \( \delta \) are constants, whose values must be carefully selected to ensure the stability and boundedness of the system over time. In this study, we adopt the parameters used in Fujii and Nakajima~\cite{inubushi2017reservoir}, and the physical constraints inherent in the modular structure further help ensure task convergence.
 
Notably, successful emulation of the NARMA-\(N\) system implies that the reservoir possesses a memory span of at least \(N\), since the nonlinearity introduced by the second and third terms requires access to the input history over \(N\) time steps.

\subsubsection*{Procedures of Readouts Training}

\begin{figure}[b!]
  \includegraphics[width=\linewidth]{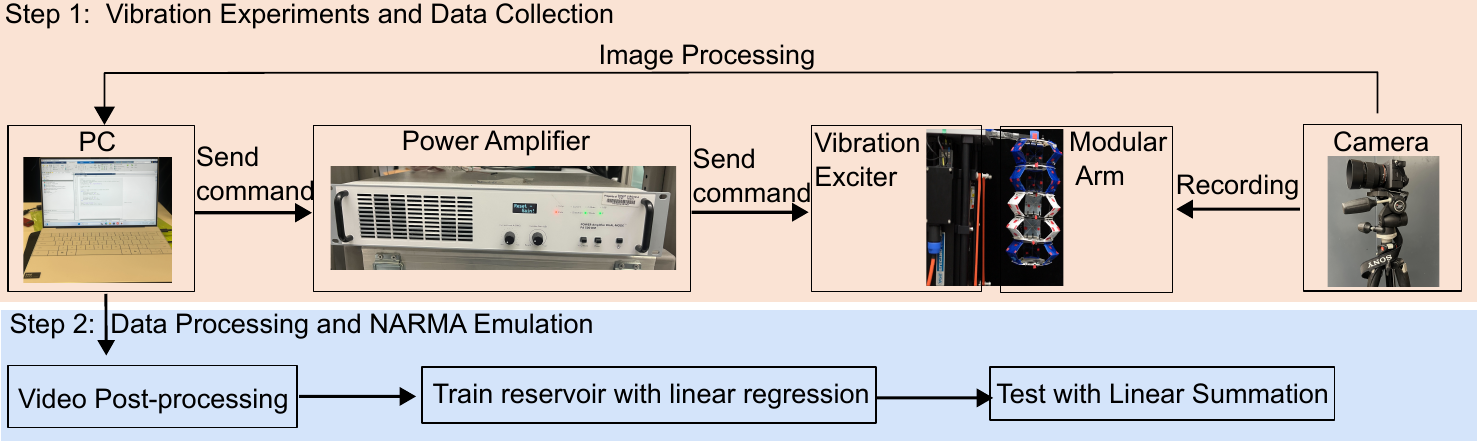}
  \centering
  \caption{Work flow chart of the vibration experiments for tasks (I) and (II)}
  \label{fig:S2}
\end{figure}

The experimental setup includes a laptop (Dell XPS 13 9340) for generating input commands, a National Instruments DAQ system (NI 9234, National Instruments, Corp.) with a power amplifier, and a large-stroke vibration exciter (APS 113, APS Dynamics, Inc.) that receives the amplified signals. A high-resolution camera (Sony a7C II) captures the displacements of all nodes/markers along the arm. The modular reservoir is connected vibration exciter, with its top end plate fixed to a rigid bar connecting to the exciter as shown in Figure~\ref{fig:S2}. In this way, the modular physical computer could receive base excitation $u(t)$ expressed as a product of three sinusoidal functions where \( (f_1, f_2, f_3) = (2.11, 3.73, 4.33) \, \text{Hz} \) times the scaling factor.

\begin{figure}[b!]
  \includegraphics[width=\linewidth]{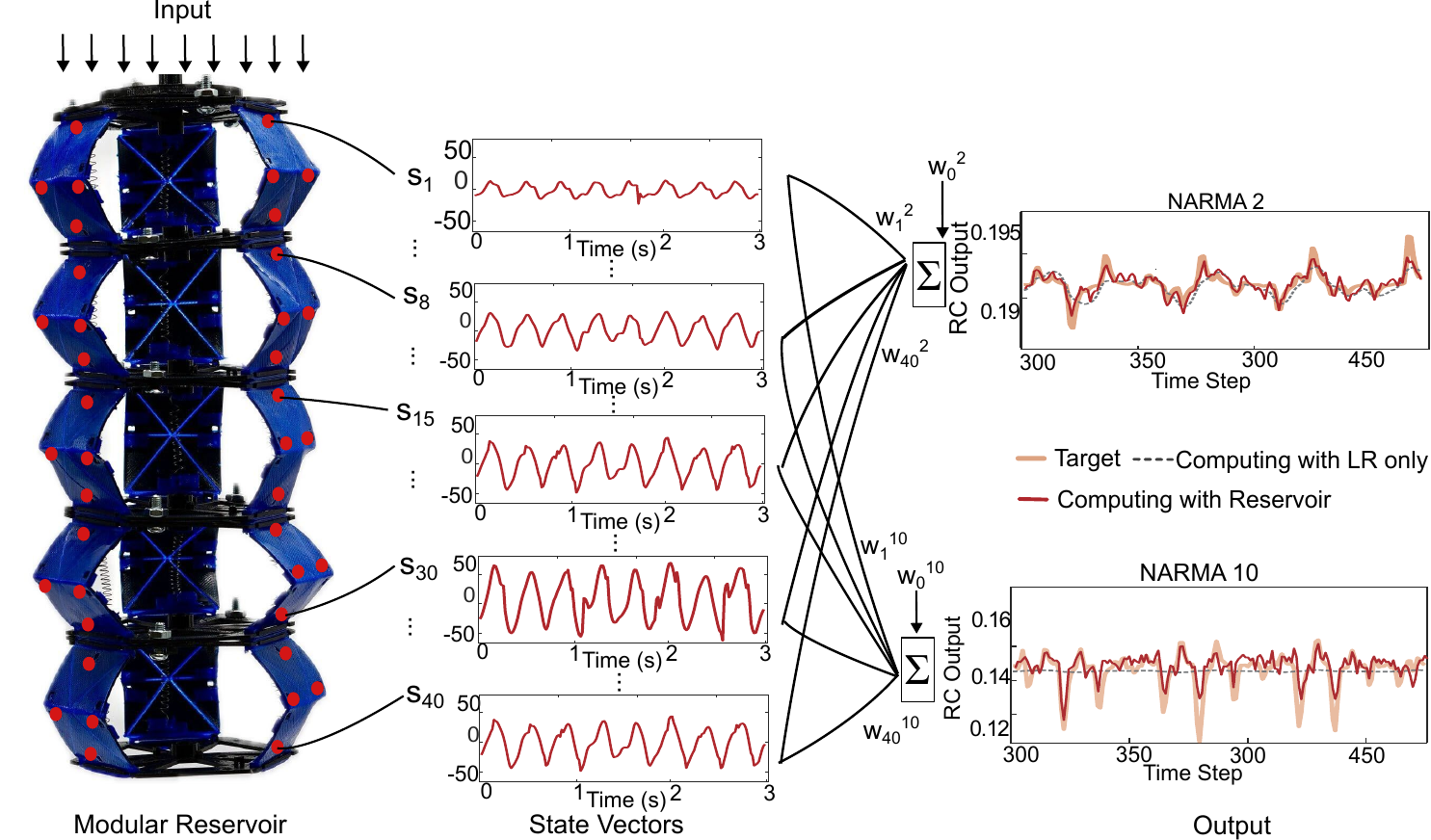}
  \centering
  \caption{Example of training the adaptive kernel at $\mathbf{C}_5$ to emulate NARMAR2 and NARMA10 in parallel. The modular structure will vibrate with the shaker after receiving an input signal from the computer. The camera records the displacement of all red markers attached to the structure. Then, readout weights for NARMA2 and NARMA10 are trained in parallel using linear regression.  }
  \label{fig:S3}
\end{figure}

Small red markers are attached to the vertices of the origami panel to track the structural dynamics. The camera records their motion at 60 frames per second with 1080p resolution. The recorded footage is post-processed in MATLAB to extract vertical displacements of all nodes as shown in Figure \ref{fig:S2}. These displacement trajectories form the reservoir state vectors $\mathbf{S}(t)=[s_1(t), s_2(t), \ldots, s_n(t)]^\intercal$, which are then used for the NARMA emulation tasks. Desired outputs \( \hat{y}^N(t) \) are obtained by training linear output weights $\mathbf{w}_\text{out}^N=[w_0^N, w_1^N, \ldots, w_7^N]^\intercal$ through linear regression such that
\begin{equation}
    \label{eq: LR}
      {\mathbf{w}_\text{out}^N}=\left[\mathbf{I} \; \mathbf{S}(t)\right]^{+}y^N(t)={\mathbf{\bar{\Phi}}(t)^{+}}y^N(t),
\end{equation}
Where $\mathbf{I} $ is a column of ones for calculating the bias term $w_0$, ${{[\cdot]}^{+}}$ is the Moore–Penrose pseudo-inverse to accommodate non-square matrices, and $\hat{\mathbf{y}}^N(t)$ is the target output, which is defined according to the task. Once the readout weights are obtained from training, the reservoir's \textit{predictions} are:

\begin{equation}
    \label{eq: output}
      \mathbf{O}^{(\text{I})}(t)=w_0+\sum\limits_{i=1}^n w_i^N s_i(t).
\end{equation}
 
Figure \ref{fig:S3} shows an example of training the adaptive kernel at configuration $\mathbf{C}_5$ to emulate NARMA2 and NARMA10 in parallel, when the input scale is set at the large $\mathbf{A}_3$. After the nodal displacements (reservoir state vectors) are extracted from the video, Equation \ref{eq: LR} is applied using  \( \hat{y}^2(t) \) and \( \hat{y}^{10}(t) \) to obtain two groups of readout weights $\mathbf{w}_\text{out}^2=[w_0^2, w_1^2, \ldots, w_n^2]^\intercal$ and $\mathbf{w}_\text{out}^{10}=[w_0^{10}, w_1^{10}, \ldots, w_n^{10}]^\intercal$. The orange line in Figure \ref{fig:S3} is the target function, and the red line is the reservoir output with trained readout weights. To explicitly show the contribution of body dynamics to the computing, we also compared the computing performance with a simple linear regression layer applied directly to the base input, \( \hat{y}^N(t) = w_0^N + w_1^N s_1(t) \), where \( w_0^N \) and \( w_1^N \) are trained only with the base node displacement from the same data set. The result of this linear regression is shown by the gray dashed line in Figure \ref{fig:S3}. All computing results with the physical body outperform those with only the linear regression layer, proving the non-negligible contribution of body dynamics to the temporal integration and nonlinearity required by the NARMA tasks.

\subsubsection*{More Details on MSE}

Experiments are conducted for each of the five structural configurations under three different input magnitudes, with each setup repeated five times, resulting in a total of 75 trials (\(5 \text{(configurations)} \times 3 \text{(input magnitude)} \times 5 \text{(repetitions)}\)). For each trial, the first 600 time steps (10 seconds) are used as a washout phase to allow transient dynamics to settle, followed by 300 time steps (5 seconds) for training, and the final 300 time steps (5 seconds) for evaluation. The errors reported in Figure~\ref{fig:RC_NARMA} represent the average NMSE across all repetitions.

To quantitatively evaluate the predictive performance of the reservoir computing system, the \textbf{Normalized Mean Square Error (NMSE)} is used. NMSE measures the deviation of the predicted output from the target values, normalized by the variance of the target signal. It is defined as:

\begin{equation}
    \text{NMSE} = \frac{\sum_{i=1}^{N}(y_i - \hat{y}_i)^2}{\sum_{i=1}^{N}(y_i - \bar{y})^2}
\end{equation}
where \( y_i \) denotes the target values, \( \hat{y}_i \) represents the predicted values from the reservoir computing model, \( \bar{y} \) is the mean of the target signal, and \( N \) is the total number of evaluation data points.

\subsubsection*{More details on PSI}

\begin{figure}[b!]
  \centering
  \includegraphics[width=\linewidth]{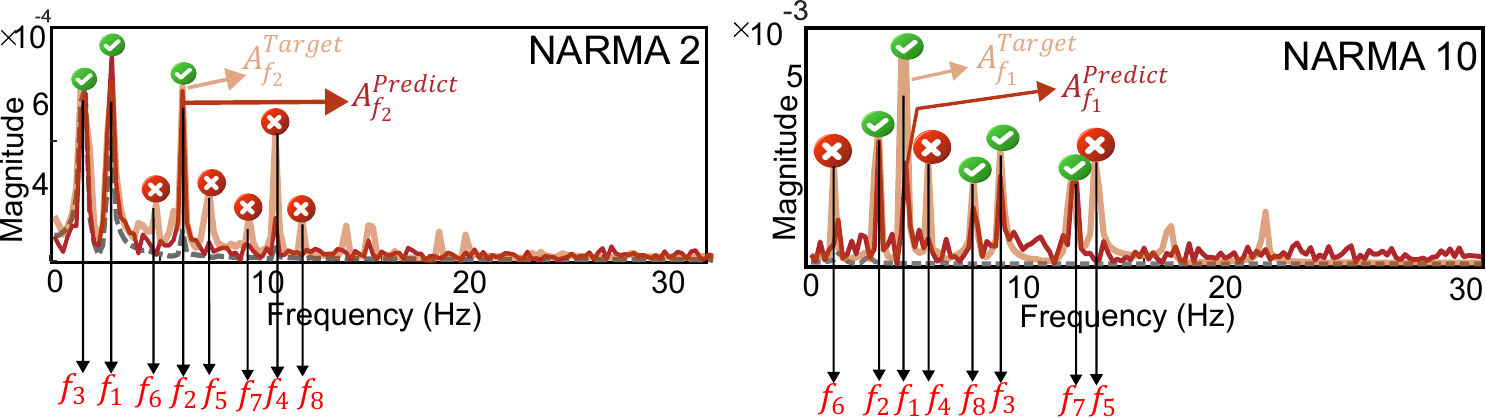}
  \caption{Spectral analysis corresponding to the emulation results of NARMA2 and NARMA10 shown in Figure~\ref{fig:S3}.}
  \label{fig:S4}
\end{figure}

Additionally, to bridge computational performance with the mechanical and frequency domains, the \textbf{Peak Similarity Index (PSI)} is introduced. PSI quantifies the discrepancy between the target and predicted signals in the frequency domain by specifically comparing the similarity of their dominant harmonic peaks. The PSI is computed in two steps. First, the magnitudes of the eight dominant harmonic peaks (\(f_1\) to \(f_8\)) are extracted from the frequency spectrum of the target signal and recorded as \(A_{f_1}^{\text{target}}\) to \(A_{f_8}^{\text{target}}\), shown as the red line in Figure~\ref{fig:S4}. Next, the magnitudes at these same frequencies are extracted from the frequency spectrum of the reservoir outputs, denoted as \(A_{f_1}^{\text{predict}}\) to \(A_{f_8}^{\text{predict}}\), shown in orange in Figure~\ref{fig:S4}.

The PSI is then calculated as:

\begin{equation}
    \text{PSI} = \sum_{i=1}^{8} \frac{A_{f_i}^{\text{predict}}}{A_{f_i}^{\text{target}}},
\end{equation}
where \(\frac{A_{f_i}^{\text{predict}}}{A_{f_i}^{\text{target}}}\) represents the similarity at the \(i\)th dominant frequency \(f_i\). A PSI value close to 8 indicates high spectral similarity between the targeted signals and reservoir output, while a value near 0 suggests the reservoir fails to capture the targeted frequency characteristics.

\begin{figure}[t!]
  \centering
  \includegraphics[width=0.6\linewidth]{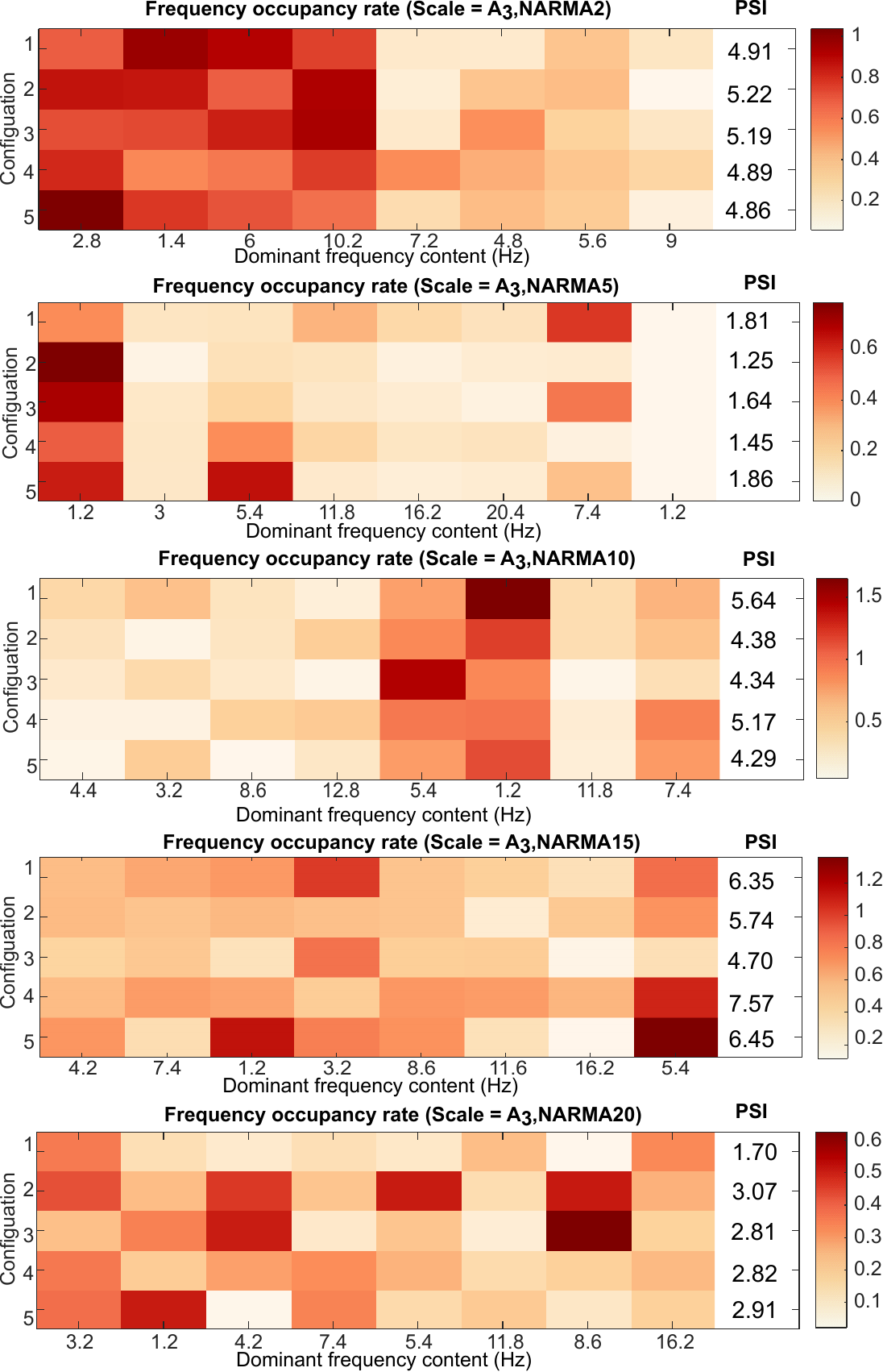}
  \caption{Frequency occupancy rates ($\frac{A_{f_i}^{\text{predict}}}{A_{f_i}^{\text{target}}}$ ) at the first eight dominant harmonic peaks of NARMA2, NARMA5, NARMA10, NARMA15, and NARMA20 tasks, evaluated using five different physical configurations. Each row in the matrix corresponds to a distinct reservoir configuration. The last column in each matrix displays the PSI value for the respective configuration, representing the overall spectral similarity between the predicted and target signals after completing the emulation task. This figure demonstrates an example of how PSI is derived, corresponding to the summary presented in Figure~\ref{fig:RC_NARMA}(b).}
  \label{fig:S5}
\end{figure}

Figure~\ref{fig:S5} summarize the $\frac{A_{f_i}^{\text{predict}}}{A_{f_i}^{\text{target}}}$ values at the first 8 dominant harmonic peaks corresponding to the five different configurations for the NARMA2, NARMA5, NARMA10, NARMA15, and NARMA20 tasks. This visualization illustrates how PSI is computed across different emulation tasks and reservoir configurations.
By directly evaluating the preservation of dominant frequency content, PSI offers an interpretable metric for assessing the ability of the physical reservoir to capture essential dynamical features, thereby linking computational accuracy with the underlying mechanical configuration.

\newpage

\subsection*{Section 3. Comparing Adaptive Reservoir to Prior Works}

We compare the NARMA performance of our adaptive kernel with that of a soft silicone arm reported in~\cite{nakajima2015information}, which has been shown to exhibit strong computational performance, even outperforming computer-simulated reservoirs in certain scenarios. In that study, the authors evaluated the NARMA tasks of order 2, 5, 10, 15, and 20 under various time scaling parameters, which modulate the phase velocity of the input time series. The minimum usable scaling factor in their experiments was 100, constrained by motor overheating at lower values.

In contrast, our system utilizes a vibration shaker for actuation, allowing for much higher input frequencies. However, our temporal resolution is constrained by the camera sampling frequency at 60\,Hz. Therefore, we compare our NARMA emulation results---based on the optimal reservoir configuration---with those from~\cite{nakajima2015information} obtained at a time scaling of 100.

The normalized mean square errors (NMSE) for NARMA tasks of orders 2, 5, 10, 15, and 20, computed using our adaptive kernel at the optimal configuration, the soft silicone body, and the artificial echo state network (ESN) reservoir in the reference study, are summarized in Table~\ref{tab:nmse_comparison}.

\begin{table}[h]
\centering
\renewcommand{\arraystretch}{1.4}
\begin{tabular}{|l|c|c|c|}
\hline
\textbf{Task} & \textbf{$\text{NMSE}^{\text{min}}_{\text{Modular}}$} & \textbf{$\text{NMSE}^{\text{system}}_{\text{Reference}}$} & \textbf{$\text{NMSE}^{\text{ESN}}_{\text{Reference}}$} \\
\hline
NARMA2  & $1.73 \pm 0.05\ (\times10^{-5})$ & $1.36 \pm 0.07\ (\times10^{-5})$ & $0.94 \pm 0.20\ (\times10^{-5})$ \\
\hline
NARMA5  & $0.69 \pm 0.04\ (\times10^{-3})$ & $1.50 \pm 0.04\ (\times10^{-3})$ & $1.72 \pm 0.90\ (\times10^{-3})$ \\
\hline
NARMA10 & $1.49 \pm 0.11\ (\times10^{-3})$ & $1.97 \pm 0.02\ (\times10^{-3})$ & $1.75 \pm 0.31\ (\times10^{-3})$ \\
\hline
NARMA15 & $3.83 \pm 0.07\ (\times10^{-3})$ & $2.81 \pm 0.03\ (\times10^{-3})$ & $2.87 \pm 0.57\ (\times10^{-3})$ \\
\hline
NARMA20 & $1.07 \pm 0.02\ (\times10^{-3})$ & $1.66 \pm 0.01\ (\times10^{-3})$ & $1.51 \pm 0.12\ (\times10^{-3})$ \\
\hline
\end{tabular}
\caption{Comparison of NMSE for different NARMA tasks using the optimized adaptive kernel, soft silicone reservoir, and artificial reservoir network from~\cite{nakajima2015information}.}
\label{tab:nmse_comparison}
\end{table}

Surprisingly, although our adaptive kernel is neither as soft as the silicone arm, nor specifically designed for emulation tasks like the artificial reservoir, its performance is comparable to or even exceeds the reference systems in NARMA5, NARMA10, and NARMA20 tasks. This result highlights that the ability to realize physical computing is not limited to specially designed soft structures but can also emerge from generic, reconfigurable, and mechanically adaptable platforms. Most importantly, it validates such versatile robotic structures for computation beyond actuation, sensing, or motion, reinforcing their potential as mechanical substrates for computation.

\newpage

\subsection{Section 4. Additional Study by Configuring a 5-segment Adaptive Kernel}

We demonstrate how to configure the adaptive kernel to maximize its computing performance through five structural configurations in the \textit{Task (I)} section, involving a different number of modules. Interestingly, even when the number of modules is fixed, the computing capability still varies depending on the specific combination of flexible and stiff states, as shown in Figure~\ref{fig:S5}. In this set of additional experiments, both the experimental setup and the training procedure follow the same protocol described in the \textit{Task I} section. 

Specifically, we fix the number of modules to five and define six distinct configurations by varying the arrangement of flexible and stiff modules. Each configuration is subjected to input signals under three different magnitude conditions, resulting in a total of \(6 \times 3 = 18\) experimental setups for the NARMA-\(N\) emulation tasks.

\begin{figure}[htbp]
  \centering
  \includegraphics[width=\linewidth]{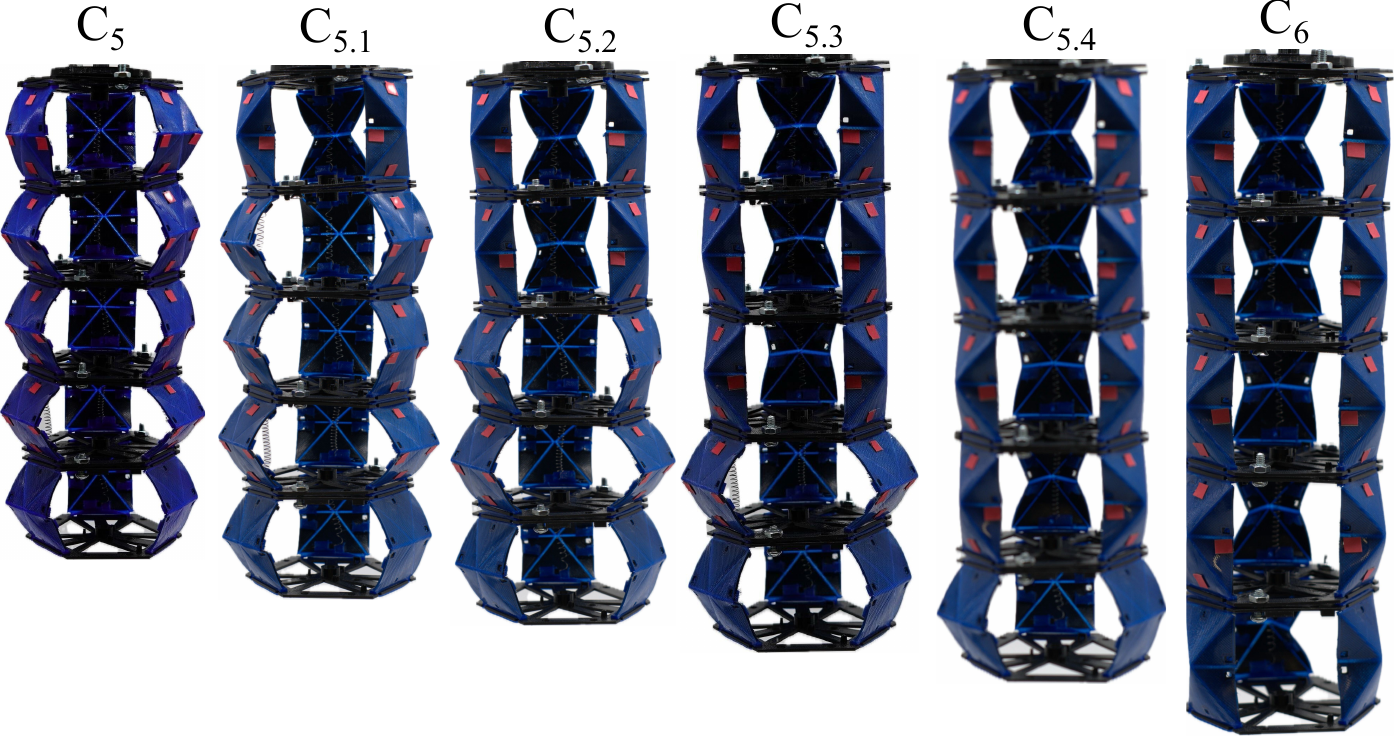}
  \caption{Six different configurations of the five-module kernel for adaptive computing. Each configuration represents a unique arrangement of flexible and stiff modules.}
  \label{fig:S5_5MODULE}
\end{figure}

Figure~\ref{fig:S5_5MODULE} summarizes the NMSE results from all 18 experimental setups across five NARMA tasks (NARMA2, 5, 10, 15, and 20). The orange bars represent the averaged NMSE for each configuration, while the overlaid line plots indicate the corresponding Peak Similarity Index (PSI). Consistent with the findings in Task (I) section in the main text, computing performance generally improves (i.e., NMSE decreases) with increased input magnitude. Furthermore, a positive correlation is observed between PSI and computational accuracy, further supporting the use of PSI as a spectral-domain performance indicator.

\begin{figure}[htbp]
  \centering
  \includegraphics[width=\linewidth]{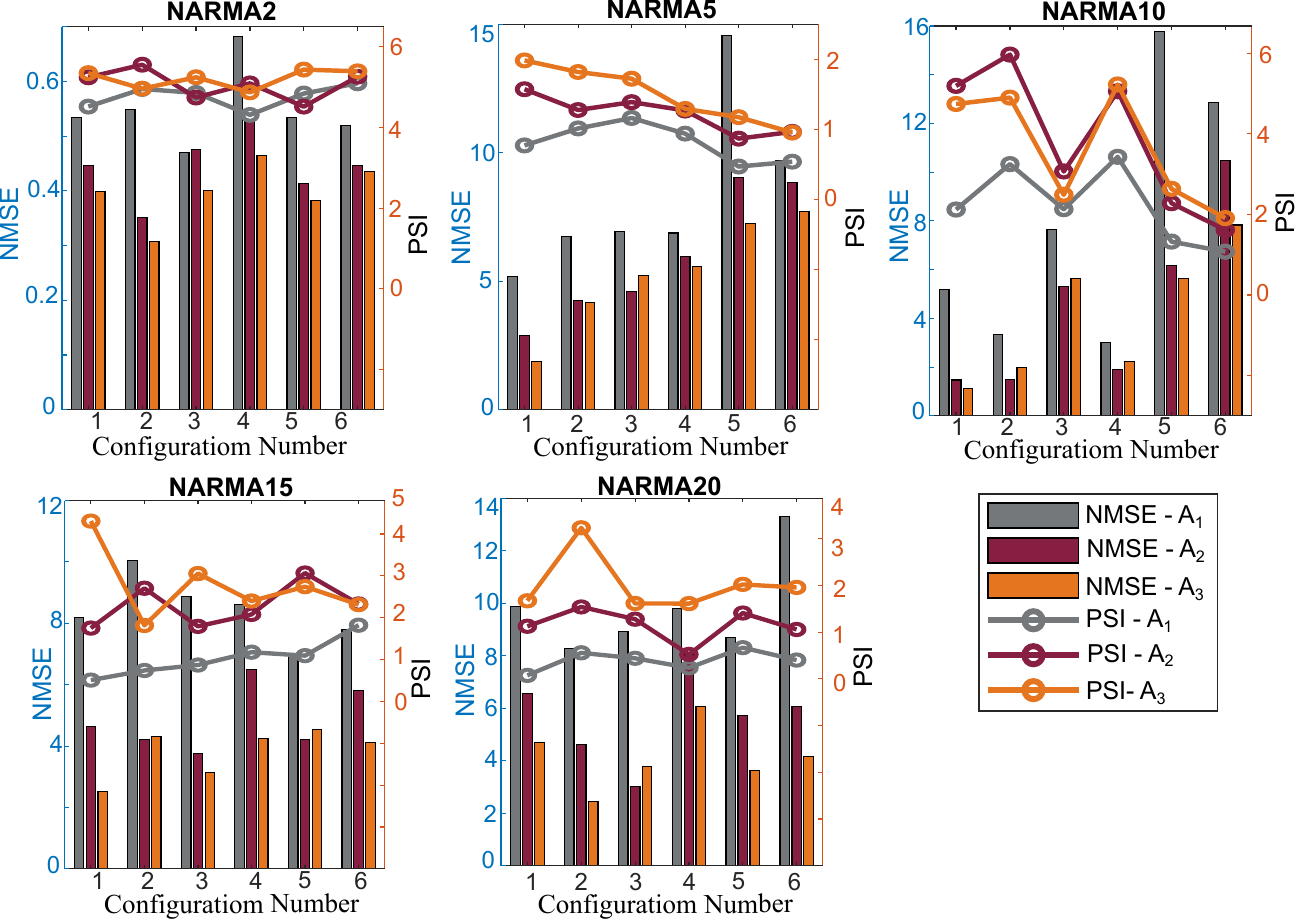}
  \caption{Averaged NMSE (bar plot) and corresponding PSI (line plot) for NARMA tasks of order 2, 5, 10, 15, and 20 across six different five-module configurations and three input magnitudes.}
  \label{fig:S6}
\end{figure}

\begin{figure}[htpb]
  \centering
  \includegraphics[width=\linewidth]{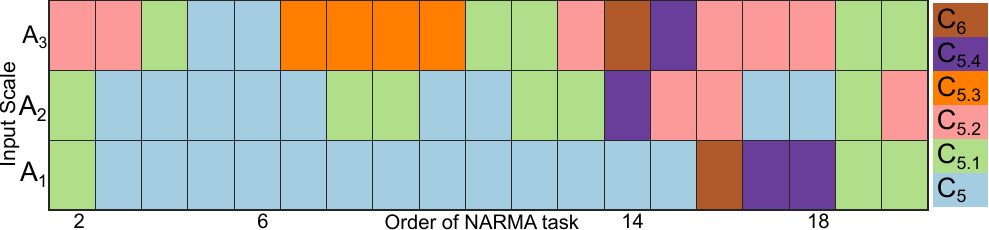}
  \caption{Optimal configuration matrix for NARMA emulation tasks of order 2 to 20 across different input magnitudes. Each cell indicates the best-performing configuration among the six tested for a given task and input scale.}
  \label{fig:S7}
\end{figure}

Finally, Figure~\ref{fig:S7} provides a comprehensive overview of the optimal configuration for NARMA emulation tasks ranging from order 2 to 20 under different input magnitudes. When the input magnitude is small, configurations with greater flexibility (i.e., the softest configuration) tend to give better computational output. In contrast, when the input magnitude increases, the optimal configuration varies across the 19 different NARMA tasks. Notably, all six configurations can serve as the optimal structure for some target functions, highlighting the adaptability and task-specific tunability of the adaptive modular reservoir.

\newpage

\subsection*{Section 5. Readout Training for Payload Mass Estimation}

\subsubsection*{Experiments Setup}

In Task (II), the experimental setup remains the same as in Task (I), as shown in Figure~\ref{fig:S19}. The adaptive kernel is connected to a shaker that provides base excitation. The only modification is that an additional payload mass (0\,g, 50\,g, 90\,g, 130\,g, or 170\,g) is attached to the last module. Consequently, the structural dynamics of the system are altered depending on the attached mass. The adaptive kernel is expected to estimate the payload mass based on its dynamic responses (i.e., through the displacements of all nodal markers).

Here, we use configuration $\mathbf{C}_5$ as an example to illustrate the process of readout training for mass estimation using different training data sizes. For each payload mass, the excitation test is repeated 10 times—one trial is used for training, and the remaining nine trials are reserved for testing. The vertical displacements of 40 nodes are extracted from the recorded videos to form the state vectors:

\begin{equation}
  \mathbf{S}_{\text{mass}}^{\text{train/test}}(t) = [s_1(t),\; s_2(t),\; \ldots,\; s_{40}(t)]^\intercal,
\end{equation}
for both training and testing. Thus, we obtain one set of training state vectors for each payload mass:

\begin{equation}
    \mathbf{S}_{0g}^{\text{train}}, \quad \mathbf{S}_{50g}^{\text{train}}, \quad \mathbf{S}_{90g}^{\text{train}}, \quad \mathbf{S}_{130g}^{\text{train}}, \quad \mathbf{S}_{170g}^{\text{train}},
\end{equation}
and nine sets of testing data:

\begin{equation}
  \mathbf{S}_{0g}^{\text{test}}, \quad \mathbf{S}_{50g}^{\text{test}}, \quad \mathbf{S}_{90g}^{\text{test}}, \quad \mathbf{S}_{130g}^{\text{test}}, \quad \mathbf{S}_{170g}^{\text{test}}.
\end{equation}

The training efficiency can be improved by reducing the number of payload mass conditions included in the training data. Specifically, Figure~\ref{fig:S10}(c) presents comprehensive results when the arm is trained using data from 1, 2, 3, 4, or all 5 payload masses. Correspondingly, the training size is reduced to 1/5, 2/5, 3/5, 4/5, and 5/5 of the full dataset, respectively.

\subsubsection*{Training with One End Mass}

\begin{figure}[htbp]
  \centering
  \includegraphics[width=\linewidth]{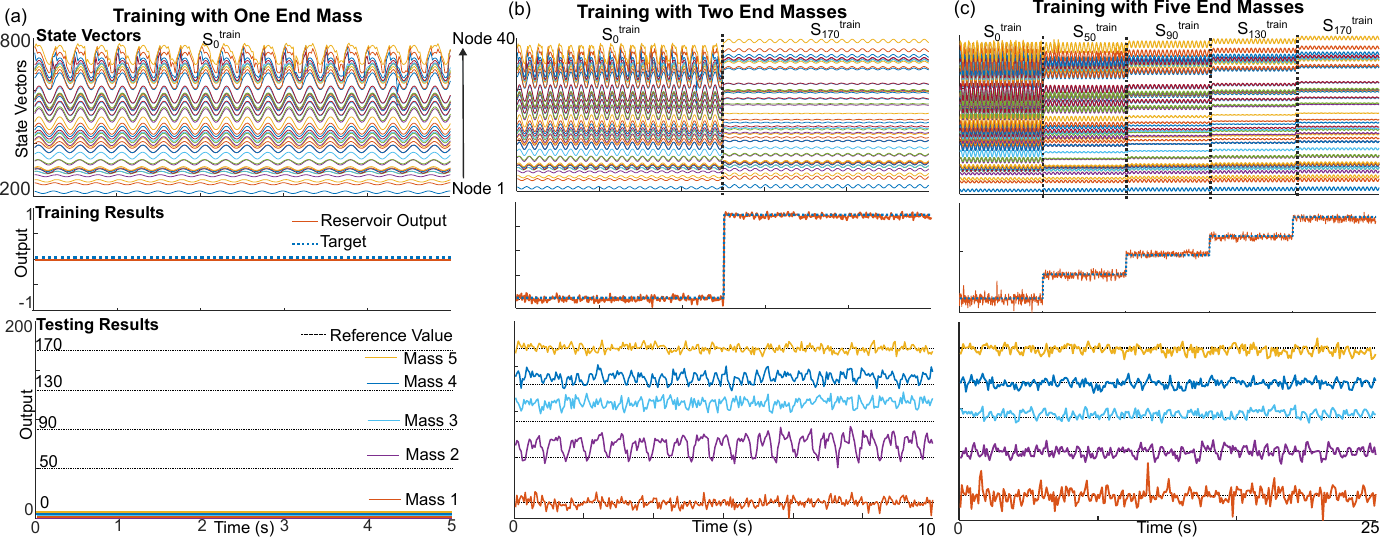}
  \caption{Training the modular reservoir $C_5$ with 4\,Hz harmonic base excitation using (a) one payload mass (0\,g), (b) two payload masses (0\,g and 170\,g), and (c) all five payload masses (0\,g, 50\,g, 90\,g, 130\,g, 170\,g). The first subplot shows the compiled state vectors used for training. The second subplot shows the corresponding target outputs and the reservoir training results. The third subplot shows the reservoir outputs for testing data using the trained readout weights.}
  \label{fig:S10}
\end{figure}

Figure~\ref{fig:S10}(a) details the results when using only a single training state vector, $\mathbf{S}_{0g}^{\text{train}}$, for readout training.

The target function for this training is set as a constant function equal to the training mass value (0\,g), represented by the blue dashed line in the second plot:

\begin{equation}
\hat{{y}}(t) = 0, \quad \text{for } 0 < t < 5\,\text{s}.    \end{equation}

A set of readout weights $\mathbf{w}_\text{out}^1 = [w_0^1, w_1^1, \ldots, w_{40}^1]^\intercal$ is trained through the linear regression Equation (\ref{eq: LR}) such that:

\begin{equation}
   \hat{{y}}(t) \approx \mathbf{w}_\text{out}^1 \mathbf{S}_{0g}^{\text{train}}(t). 
\end{equation}

The second subplot of Figure \ref{fig:S10} illustrates the good agreement between the reservoir output (orange curve) and the target function (blue dashed line).

After training, the obtained readout weights $\mathbf{w}_\text{out}^1 = [w_0^1, w_1^1, \ldots, w_{40}^1]^\intercal$ are applied to the testing datasets to evaluate estimation performance. The colorful lines in the third subplot represent the testing outputs:

\begin{equation}
    {y}_{0g}^{\text{test}} = \mathbf{w}_\text{out}^1 \mathbf{S}_{0g}^{\text{test}}, \quad {y}_{50g}^{\text{test}} = \mathbf{w}_\text{out}^1 \mathbf{S}_{50g}^{\text{test}}, \quad {y}_{90g}^{\text{test}} = \mathbf{w}_\text{out}^1 \mathbf{S}_{90g}^{\text{test}}, \quad
    {y}_{130g}^{\text{test}} = \mathbf{w}_\text{out}^1 \mathbf{S}_{130g}^{\text{test}}, \quad {y}_{170g}^{\text{test}} = \mathbf{w}_\text{out}^1 \mathbf{S}_{170g}^{\text{test}}. 
\end{equation}

The estimated mass value $\bar{y}_{\text{out}}$ is computed by averaging the reservoir outputs over all time steps (5\,s) as:

\begin{equation}
    \bar{y}_{\text{out}} = \frac{1}{T} \sum_{t=1}^{T} y_{\text{out}}(t).
\end{equation}

The results show that the estimated outputs for all different payload masses are close to zero (far from the ground truth), demonstrating that training with only one mass condition is insufficient for accurate mass estimation across different payloads.

\subsubsection*{Training with Two End Masses}

We increase the training data by using the nodal displacement from two end masses to improve the reservoir's performance. Specifically, two training state vectors corresponding to the two extreme payload conditions---no payload (0\,g) and maximum payload (170\,g)---are compiled for training, as shown in Figure~\ref{fig:S10}(b). The training vectors are concatenated in time to form a single sequence:

\begin{equation}
    \mathbf{S}^{\text{train}} = 
      \begin{bmatrix}
        \mathbf{S}^{\text{train}}_{0\,\text{g}} \\
        \mathbf{S}^{\text{train}}_{170\,\text{g}}
      \end{bmatrix},
\end{equation}

as depicted in the top row of each subplot in Figure~\ref{fig:S10}. The corresponding target function is defined as:

\begin{equation}
    \hat{{y}}(t) = 
        \begin{cases}
          0, & \text{for } 0 < t < 5\,\text{s}, \\
        170, & \text{for } 5 < t < 10\,\text{s}.
        \end{cases}
\end{equation}

Notably, the total training duration is extended from 5\,s to 10\,s. A new set of readout weights $\mathbf{w}_\text{out}^2 = [w_0^2, w_1^2, \ldots, w_{40}^2]^\intercal$ is obtained by applying the linear regression Equation (\ref{eq: LR}) to the concatenated training data, resulting in:

\begin{equation}
    \hat{{y}}(t) \approx \mathbf{w}_\text{out}^2 \mathbf{S}^{\text{train}}(t).
\end{equation}

The second subplot in Figure~\ref{fig:S10} demonstrates the close alignment between the reservoir output (orange curve) and the target function (blue dashed line). 

The testing process remains the same: the newly trained readout weights are applied to the testing datasets. The third subplot in Figure~\ref{fig:S10} shows the reservoir outputs for each testing condition, namely ${y}_{0g}^{\text{test}}$, ${y}_{50g}^{\text{test}}$, ${y}_{90g}^{\text{test}}$, ${y}_{130g}^{\text{test}}$, and ${y}_{170g}^{\text{test}}$. 

Compared to the case trained with only one mass, the adaptive kernel now demonstrates significantly improved separability between different payloads, with only slight deviations between the target and actual outputs.

\subsubsection*{Training with Five End Masses}

For the most comprehensive training scenario, we use the nodal displacement from all five available payload masses---0\,g, 50\,g, 90\,g, 130\,g, and 170\,g---for readout training. The corresponding state vectors are concatenated sequentially to form the full training sequence:

\begin{equation}
    \mathbf{S}^{\text{train}} = 
     \begin{bmatrix}
        \mathbf{S}^{\text{train}}_{0\,\text{g}} \\
        \mathbf{S}^{\text{train}}_{50\,\text{g}} \\
        \mathbf{S}^{\text{train}}_{90\,\text{g}} \\
        \mathbf{S}^{\text{train}}_{130\,\text{g}} \\
        \mathbf{S}^{\text{train}}_{170\,\text{g}}
      \end{bmatrix}.
\end{equation}

The corresponding target function is defined as a piecewise-constant signal matching the payload weight for each time interval:

\begin{equation}
    \hat{{y}}(t) = 
      \begin{cases}
        0, & \text{for } 0 < t < 5\,\text{s}, \\
        50, & \text{for } 5 < t < 10\,\text{s}, \\
        90, & \text{for } 10 < t < 15\,\text{s}, \\
        130, & \text{for } 15 < t < 20\,\text{s}, \\
        170, & \text{for } 20 < t < 25\,\text{s}.
       \end{cases}
\end{equation}

Thus, the total training duration increases to 25\,s. A new set of readout weights $\mathbf{w}_\text{out}^5 = [w_0^5, w_1^5, \ldots, w_{40}^5]^\intercal$ is trained using the linear regression Equation (\ref{eq: LR}), resulting in:

\begin{equation}
    \hat{{y}}(t) \approx \mathbf{w}_\text{out}^5 \mathbf{S}^{\text{train}}(t).
\end{equation}

As shown in Figure~\ref{fig:S10}(c), the reservoir output closely follows the target function during training, confirming that the system successfully encodes the payload-dependent dynamics.

The same trained readout weights are then applied to the testing datasets. The third subplot in Figure~\ref{fig:S10} shows the testing results for all payload conditions. The outputs exhibit clear separability across different masses, with minimal deviations from the true mass values. This result indicates that using a broader range of training conditions enhances the reservoir’s generalization ability and improves mass estimation accuracy across unseen payloads.

\subsubsection*{Definition of spatial correlation}

The displacement data from 40 nodes distributed along the adaptive kernel's body are recorded to characterize its dynamic properties. To quantitatively describe the relationships between these nodes, a correlation matrix \( \mathbf{C}_\text{orr} \) is constructed, where each element \( C_{\text{orr},ij} \) represents the correlation coefficient between the displacement time series of node \( i \) and node \( j \). The correlation matrix is formally defined as:

\begin{equation}
   \mathbf{C}_\text{orr} = 
\begin{bmatrix}
C_{\text{orr},11} & C_{\text{orr},12} & \dots & C_{\text{orr},1N} \\
C_{\text{orr},21} & C_{\text{orr}, 22} & \dots & C_{\text{orr},2N} \\
\vdots & \vdots & \ddots & \vdots \\
C_{\text{orr},N1} & C_{\text{orr},N2} & \dots & C_{\text{orr},NN}
\end{bmatrix} 
\end{equation}

where \( N = 40 \), and each entry \( C_{\text{orr},ij} \) is calculated as:

\begin{equation}
    C_{\text{orr}, ij} = \frac{\mathrm{cov}(s_i, s_j)}{\sigma_{s_i}\sigma_{s_j}}.
\end{equation}

Here, \( s_i \) and \( s_j \) denote the displacement data from nodes \( i \) and \( j \), respectively, \( \mathrm{cov}(s_i,s_j) \) is their covariance, and \( \sigma_{s_i} \), \( \sigma_{s_j} \) are the standard deviations.

The correlation index \( R_i \) for each node \( i \) is then defined as the sum of correlations between node \( i \) and all other nodes in the structure, given by:

\begin{equation}
    R_i = \sum_{j=1}^{N} C_{\text{orr},ij}.
\end{equation}

The average correlation index of the whole structure, Avg. CI, is then defined as the average of the correlation index for all nodes.

\begin{equation}
    Avg. CI =\frac{ \sum_{i=1}^{N} R_{i}}{N}
\end{equation}

The correlation index provides an insightful measure indicating each node's average connectivity or interaction strength with the overall structural dynamics, thus enabling the assessment of nodes with significant influence or sensitivity within the structure.

\newpage

\subsection*{Section 6. Setup and Readout Training for the Adaptive Robotic Kernel}

\subsubsection*{More details about the SMA actuation setup, mechatronic diagrams}

\begin{figure}[htbp]
  \includegraphics[width=0.9\linewidth]{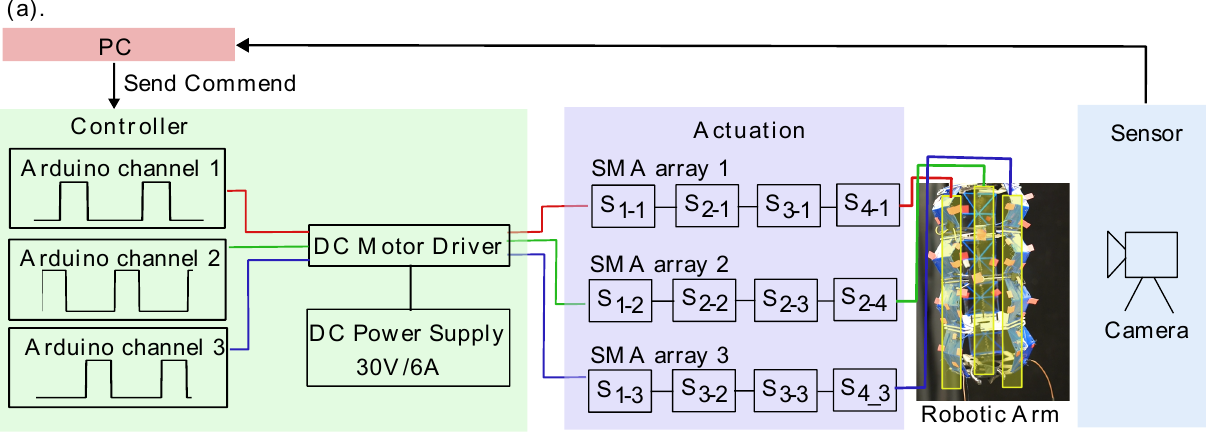}
  \centering
  \caption{Mechatronic design for actuating the modular arm.}
  \label{fig:S13}
\end{figure}

The robotic kernel designed for manipulation tasks features a four-module setup, as introduced in Section Task (III). The Shape Memory Alloys (SMAs) used in this task are Nitinol helical springs (Kellogg's Research Labs, wire diameter: 0.5mm, mandrel diameter: 4.75mm, transition temperature: 45\degree C). Each spring is cut into twelve segments, resulting in SMA coils of 20mm in length, which are then integrated into the modular robotic system to enable thermally-induced actuation. Each Shape Memory Alloy (SMA) embedded in these modules is labeled as \( S_{i-j} \), where \( i \) (with values 1, 2, 3, or 4) indicates the module number, and \( j \) (with values 1, 2, or 3) denotes the column of the SMA. In total, there are four SMAs arranged in each column, which are connected in series. This setup includes three arrays of SMAs, as depicted in Figure \ref{fig:S13}.

Each SMA array is responsible for inducing motion in one bending direction of the arm. These arrays are activated sequentially using a PWM signal generated by an Arduino Uno. The mechatronics diagram controlling the robotic arm is illustrated in Figure \ref{fig:S13}(a). The arm operates on a control logic that cycles through each SMA array, with an activation period of 0.5 seconds and a total cycle time of 1.5 seconds. The Arduino outputs PWM signals to three channels linked to one SMA through an MDD10A dual-channel motor driver. Since the MDD10A only supports two channels, two DC motors drivers are included in the setup. A 6V power supply, specifically the BK Precision 9131C rated at 5A, provides ample current to heat the SMAs to their transformation temperature quickly.

During each actuation cycle, SMA1 is activated first, followed by SMA2 and SMA3, allowing the robotic arm to bend sequentially at a high frequency without the risk of overheating. This configuration enables the robotic system to exhibit dynamic and cyclic motion patterns with minimal mechanical complexity, effectively utilizing the intrinsic dynamics of the whole structure for input command reconstruction and payload information identification.


The controlling command of Arduino is given by: 
\begin{lstlisting}[style=arduino, caption={Arduino code for SMA sequential control}]
const int sma1Pin = 3;  // PWM pin for SMA1
const int sma2Pin = 5;  // PWM pin for SMA2
const int sma3Pin = 6;  // PWM pin for SMA3

const int pwmValue = 204; // ~80% duty cycle
const unsigned long onTime = 500;  // 0.5 seconds
const unsigned long offTime = 1000; // Wait time between activations

void setup() {
  pinMode(sma1Pin, OUTPUT);
  pinMode(sma2Pin, OUTPUT);
  pinMode(sma3Pin, OUTPUT);
}

void loop() {
  analogWrite(sma1Pin, pwmValue);
  analogWrite(sma2Pin, 0);
  analogWrite(sma3Pin, 0);
  delay(onTime);
  analogWrite(sma1Pin, 0);
  delay(onTime);
  analogWrite(sma2Pin, pwmValue);
  delay(onTime);
  analogWrite(sma2Pin, 0);
  delay(onTime);
  analogWrite(sma3Pin, pwmValue);
  delay(onTime);
  analogWrite(sma3Pin, 0);
  delay(onTime);
}
\end{lstlisting}

\subsubsection*{Readout Training For SMA Input Reconstruction}

\begin{figure}[htbp]
  \centering
  \includegraphics[width=0.9\linewidth]{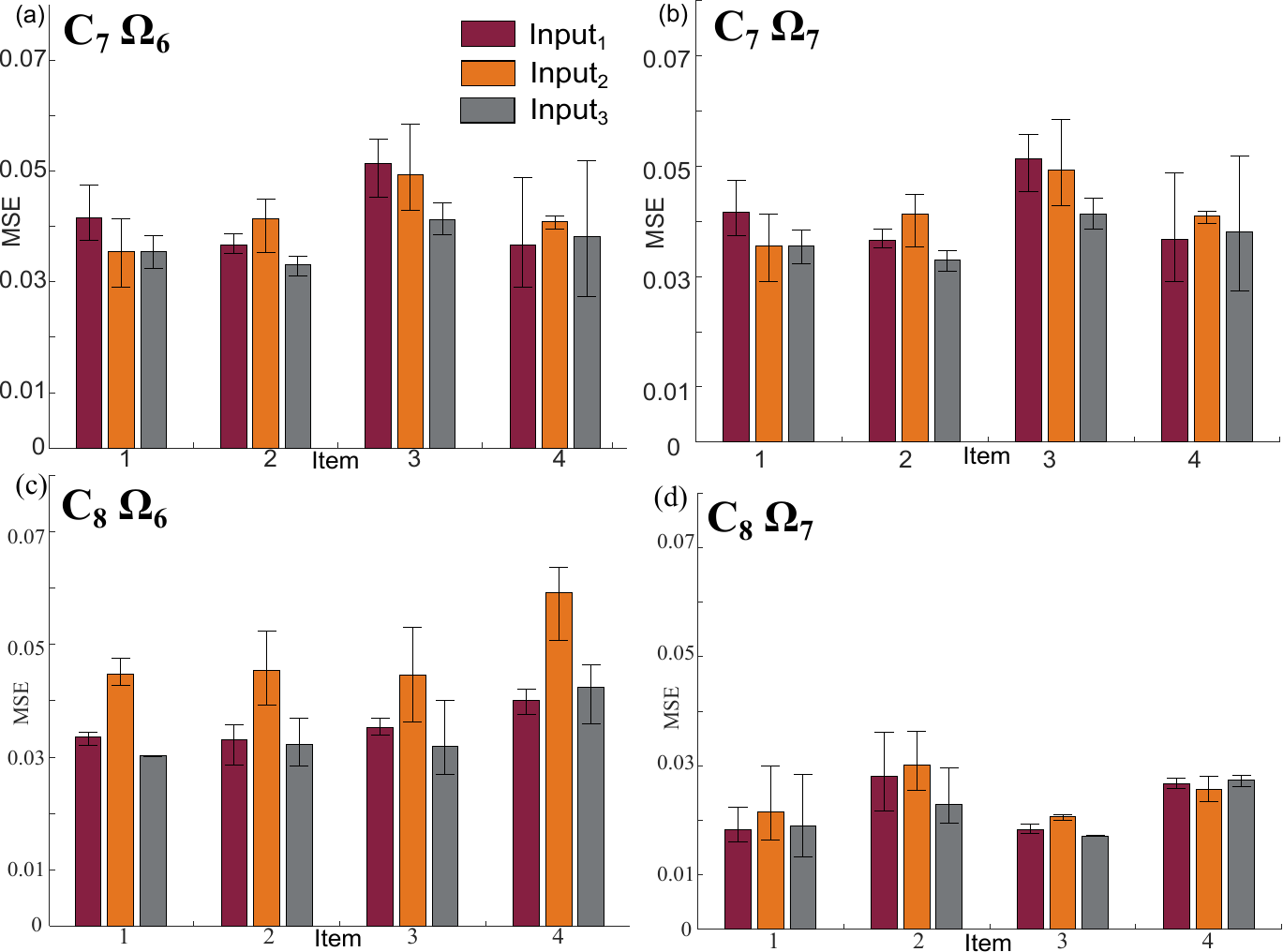}
  \caption{MSE of input reconstruction tasks. (a) shows the MSE bar plot when the arm in configuration $C_7$ is actuated at frequency $f_6$ under four different payload conditions. Similarly, (b), (c), and (d) present the results for $C_7f_7$, $C_8f_6$, and $C_8f_7$, respectively. Each bar represents the average MSE over 10 trials, and the error bars indicate the minimum and maximum values.}
  \label{fig:S17}
\end{figure}

We attach four items to the robotic kernel's free end, one at a time, to train the readouts for the input reconstruction task (Figure~\ref{fig:S17}). After the SMAs were heated to their transition temperature, the robotic arm initially exhibited large deformations toward the left-front and right-front directions during the transient phase. It then swung slightly around a balanced bending position oriented toward the front. This swinging motion was induced by the alternating actuation of SMA arrays 1, 2, and 3. The dynamic responses, represented by red marker's displacement distributed along the kernel, were captured by a camera and used to construct the state vectors $\mathbf{S}(t) = [s_1(t), s_2(t), \ldots, s_{40}(t)]^\intercal$ for readout training.

\begin{figure}[htbp]
  \centering
  \includegraphics[width=0.8\linewidth]{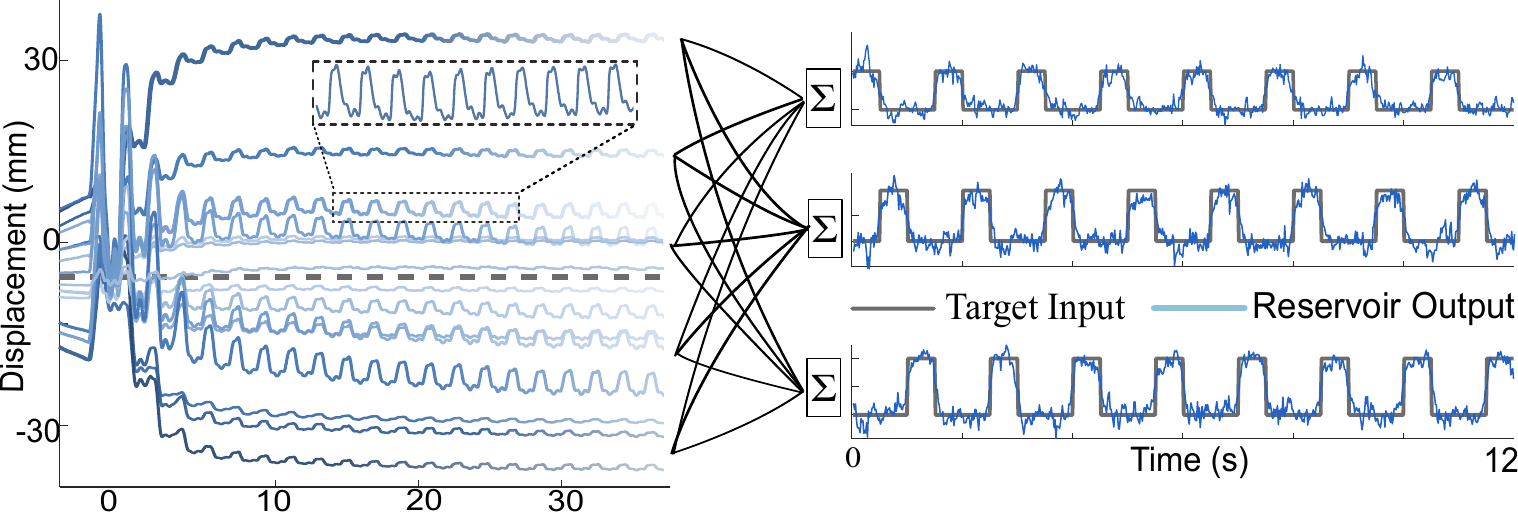}
  \caption{Training process for three SMA command emulations using displacement data collected when configuration $C_7$ is actuated at frequency $f_8$ (1.5\,s per cycle).}
  \label{fig:S15}
\end{figure}

The three input PWM signals generated by the Arduino, shown in gray in Figure~\ref{fig:S15}, were used as target functions ${y}$:

\begin{equation}
    {{y}}_i(t) = \text{Input}_i \quad (i = 1, 2, 3),
\end{equation}

and three distinct sets of readout weights, $\mathbf{w}_\text{out}^1$, $\mathbf{w}_\text{out}^2$, and $\mathbf{w}_\text{out}^3$, were trained in parallel using Equation (\ref{eq: LR}). The reservoir outputs during the training phase are shown in blue in Figure~\ref{fig:S15}, illustrating the system’s ability to rebuild the target inputs.

To evaluate the reconstruction performance, we computed the Mean Squared Error (MSE) between the predicted signals and the ground truth inputs during the testing phase:

\begin{equation}
    \text{MSE}_i = \frac{1}{T} \sum_{t=1}^{T} \left( \hat{y}_3^i(t) - y_3^i(t) \right)^2, \quad i = 1, 2, 3.
\end{equation}

In each experimental condition (i.e., each payload setup), the robotic kernel was actuated for 30 seconds, and the experiment was repeated 10 times. For each trial, the first 15 seconds were used for training, and the remaining 15 seconds were used for testing. The Mean Squared Error (MSE) was calculated over the testing interval of each trial. In the bar plots shown in Figure~\ref{fig:S17}, the average MSE across the 10 trials is reported for each condition, and the error bars indicate the minimum and maximum MSE values observed. This evaluation protocol captures the variability of the system under repeated actuation and demonstrates the robustness of its reconstruction capability across different dynamic regimes and load conditions.

Overall, the results suggest that higher-frequency actuation ($\mathbf{\Omega}_7$) is more beneficial for proprioceptive information estimation. This could be attributed to the fact that, under higher-frequency stimulation, the robotic kernel's vibration becomes more synchronized with the input command, thereby enabling more accurate regeneration of the original control signal. Furthermore, it is observed that the reconstruction accuracy is notably higher for input commands applied to columns 1 and 3 when the arm is in configuration $\mathbf{C}_8$. This is likely because column 2 remains stiff (state '1') and cannot move, introducing asymmetry in the actuation pattern. However, this immobility stabilizes the overall bending behavior, contributing to consistent dynamic responses that generally make $\mathbf{C}_8$ outperform $\mathbf{C}_7$.

\subsubsection*{Readout Training for Payload Weight and Orientation Classification}

For the payload perception task using the SMA-actuated robotic kernel, we use different workshop tools, such as pliers, screwdrivers, and hammers, as perception targets to bring the task closer to practical applications. The objective is for the reservoir-enabled robotic arm to not only classify the object being held but also infer its orientations---for instance, determining whether a hammer is oriented with its head towards the front, left, or right. This form of exteroception is achieved through a two-step learning procedure: (1) payload weight estimation for coarse classification and (2) orientation classification for finer-grained recognition. The arm can infer both the object type and its spatial configuration because its dynamic response is influenced by the payload’s weight and the centroid location relative to the grasping point.

\begin{figure}[htbp]
  \centering
  \includegraphics[width=\linewidth]{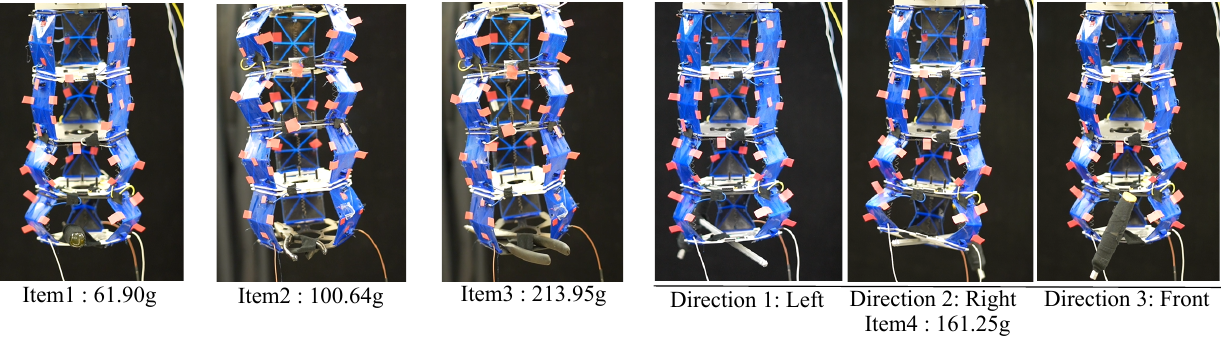}
  \caption{Initial setup showing the modular arm holding four different items. Item 4 (hammer) is additionally tested in three orientations (front, left, and right).}
  \label{fig:S16}
\end{figure}

\begin{figure}[htbp]
  \centering
  \includegraphics[width=0.9\linewidth]{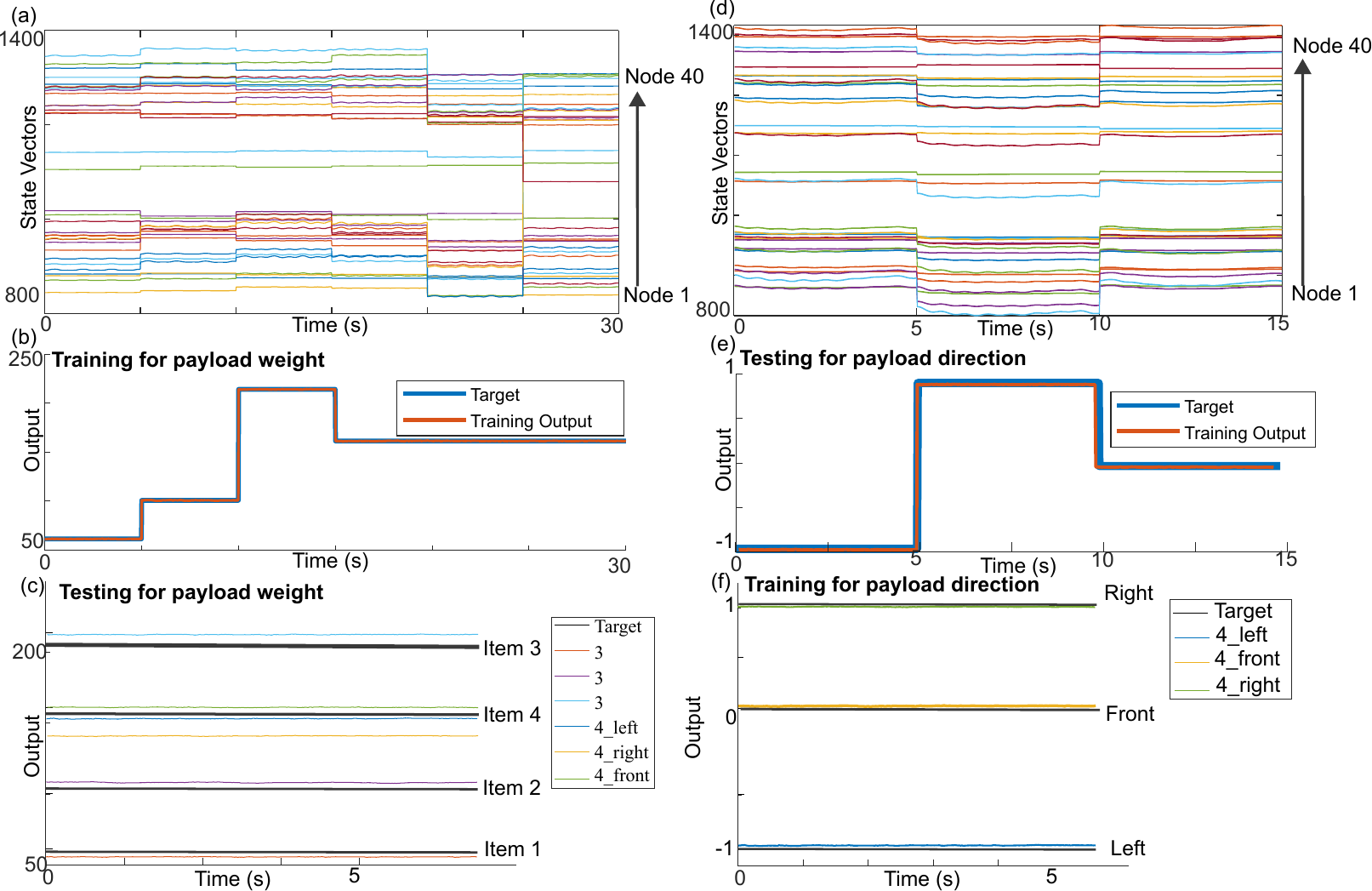}
  \caption{(a-c). Training and testing results for the payload weight estimation task. (d-f). Training and testing results for the hammer orientation classification task.}
  \label{fig:S18}
\end{figure}

The arm is trained to perform payload weight estimation in the first step, following the method in the previous Task (II). For each of the six payload scenarios shown in Figure~\ref{fig:S16}, 5 seconds of displacement data from all red markers along the arm are recorded. These sequences are concatenated to form a single state vector:

\begin{equation}
    \mathbf{S}(t) = 
\begin{bmatrix}
\mathbf{S}^{\text{item}_1}(t) \\
\mathbf{S}^{\text{item}_2}(t) \\
\mathbf{S}^{\text{item}_3}(t) \\
\mathbf{S}^{\text{item}_{4L}}(t) \\
\mathbf{S}^{\text{item}_{4R}}(t) \\
\mathbf{S}^{\text{item}_{4F}}(t)
\end{bmatrix},
\end{equation}

as illustrated in Figure~\ref{fig:S18}(a). The corresponding target function is defined to represent the estimated effective weight of each payload:

\begin{equation}
    \hat{\mathbf{y}}_{4}(t) = 
\begin{cases}
61.90, & \text{for } 0 < t < 5\,\text{s}, \\
100.64, & \text{for } 5 < t < 10\,\text{s}, \\
213.95, & \text{for } 10 < t < 15\,\text{s}, \\
161.25, & \text{for } 15 < t < 20\,\text{s}, \\
161.25, & \text{for } 20 < t < 25\,\text{s}, \\
161.25, & \text{for } 25 < t < 30\,\text{s}.
\end{cases}
\end{equation}

The readout weights $\mathbf{w}_\text{out}^6$ are trained using these input–output pairs. As shown in Figure~\ref{fig:S18}(b), the reservoir output aligns well with the target function, demonstrating effective encoding of weight-based classification.

\begin{figure}[htbp]
  \centering
  \includegraphics[width=0.8\linewidth]{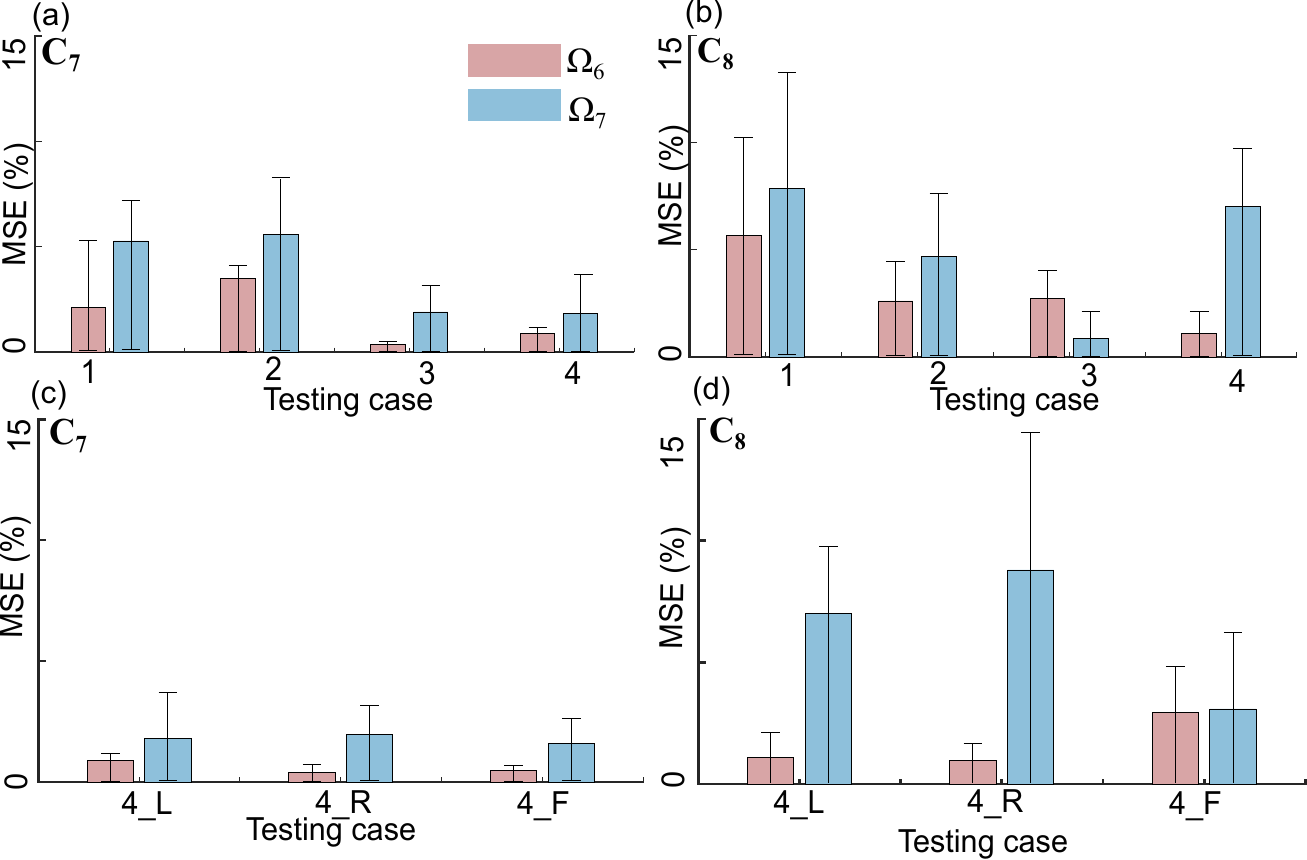}
  \caption{MSE results for the payload perception tasks. (a, b) MSE of weight estimation under configurations $C_7$ and $C_8$. (c, d) MSE of orientation classification under the same configurations.}
  \label{fig:S19}
\end{figure}

After training, each payload condition was tested in 10 repeated trials, each lasting 30 seconds. This results in $6 \times 10$ groups of state vectors. During testing, the reservoir outputs were computed using the pre-trained readout weights via ${y}_{4}(t) = \mathbf{w}_\text{out}^6 \mathbf{S}_\text{test}(t)$. The Mean Squared Error (MSE) was computed over each testing interval. As shown in Figure~\ref{fig:S19}(a) and (b), the average MSE across all trials remains below 5\% for both robotic configurations $\mathbf{C}_7$ and $\mathbf{C}_8$. Notably, lower actuation frequency (0.33\,Hz) resulted in significantly smaller variance.

Building on the weight estimation task results, we further train the robotic arm kernel to infer the hammer’s orientation, provided that the average predicted weight lies between 140\,g and 180\,g---indicating that item 4 (the hammer) is being held. We extract displacement data corresponding to the three orientations (left, front, right) of the hammer, as shown in Figure~\ref{fig:S16}. The sequences are concatenated to form the orientation-specific training input as shown in Figure~\ref{fig:S18}(d):

\begin{equation}
    \mathbf{S}(t) = 
\begin{bmatrix}
\mathbf{S}^{\text{item}_{4L}}(t) \\
\mathbf{S}^{\text{item}_{4F}}(t) \\
\mathbf{S}^{\text{item}_{4R}}(t)
\end{bmatrix}.
\end{equation}

This task is treated as a classification problem with three discrete direction labels. The target function for direction perception is defined as:

\begin{equation}
    \label{eq:detect_target}
    \hat{\mathbf{y}}_{5}(t) =
    \begin{cases}
      -1, & \text{if the hammer is heading left}, \\
       0, & \text{if the hammer is heading front}, \\
       1, & \text{if the hammer is heading right}.
    \end{cases}
\end{equation}

The readout weights $\mathbf{w}_\text{out}^7$ are trained using these labels and corresponding state vectors. The training results shown in Figure~\ref{fig:S18}(e) confirm that the reservoir successfully captures directional information encoded in the dynamic response.

Testing is conducted over 10 repeated trials for each hammer orientation. The MSE of classification results is reported in Figure~\ref{fig:S19}(c) and (d). Across both configurations, the system reliably distinguishes between different orientations with low variance, particularly under configuration $\mathbf{C}_7$. Additionally, slower actuation stabilizes the dynamic profile and reduces prediction variance. This implies that lower-frequency motion leads to richer, more separable dynamics for exteroceptive tasks such as orientation inference.

\end{document}